%% file: cvpr15.tex
\PassOptionsToPackage{usenames,dvipsnames,table}{xcolor}
\documentclass[10pt,twocolumn,letterpaper]{article}

\usepackage{cvpr}
\usepackage{times}
\usepackage{epsfig}
\usepackage{graphicx}
\usepackage{amsmath}
\usepackage{amssymb}
\usepackage{booktabs}
\usepackage{multirow}

\usepackage{authblk}
\usepackage[inline]{enumitem}

\cvprfinalcopy 

\def\cvprPaperID{****} 

\usepackage{amsmath}
\usepackage{booktabs}

\usepackage[pagebackref=true,breaklinks=true,letterpaper=true,colorlinks,bookmarks=false]{hyperref}

\def\CNN{ConvNet}
\def\CNNs{ConvNets}
\def\alexnet{\texttt{AlexNet}}
\def\caffe{\texttt{Caffe}}
\def\bff{\mathbf{f}}
\def\by{\mathbf{y}}
\def\bx{\mathbf{x}}
\def\bw{\mathbf{w}}
\def\rep{representation}

\graphicspath{{./figures/}}

\begin{document}

\title{Persistent Evidence of Local Image Properties in Generic ConvNets}


\author{Ali Sharif Razavian}
\author{Hossein Azizpour}
\author{\\\vspace{-0.2cm}Atsuto Maki}
\author{Josephine Sullivan}
\author{Carl Henrik Ek}
\author{Stefan Carlsson}

\affil{CVAP, KTH (Royal Institute of Technology), Stockholm,
  SE-10044\\
{\tt\small
  \{razavian,azizpour,atsuto,sullivan,chek,stefanc\}@csc.kth.se}  
}

\maketitle

\begin{abstract}
\input{abstract.tex}

\end{abstract}


\input{Introduction.tex}

\input{Motivation.tex}
\input{Experiment.tex}

\input{Feature_Extrapolation.tex}

\input{Conclusion.tex}

\section*{Acknowledgment}
We would like to gratefully acknowledge the support of NVIDIA for the donation of multiple GPU cards for this research.

{\small
\bibliographystyle{ieee}
\bibliography{egbib}
}

\end{document}

%% file: abstract.tex
Supervised training of a convolutional network for object classification
should make explicit any information related to the class of objects and disregard
any auxiliary information associated with the capture of the image or the
variation within the object class. Does this happen in practice? Although this seems to pertain to the very final
layers in the network, if we look at earlier layers we find that this is not the
case. Surprisingly, strong spatial information is implicit.
This paper addresses this, in particular, exploiting the image representation 
at the first fully connected layer, i.e. the global image descriptor 
which has been recently shown to be most effective in a range of visual recognition tasks. 
We empirically demonstrate evidences for the finding in the contexts of 
four different tasks: 2d landmark detection, 2d object keypoints prediction, 
estimation of the RGB values of input image, and recovery of semantic 
label of each pixel. We base our investigation on a simple framework with 
ridge rigression commonly across these tasks, and show results which all support 
our insight. Such spatial information can be used for computing correspondence of
landmarks to a good accuracy, but should potentially be useful for improving
the training of the convolutional nets for classification purposes.

%% file: Introduction.tex
\section{Introduction}
\begin{figure}
  \centering
  \begin{tabular}{@{}cc@{}}
    \multirow{2}{*}[9.5em]{
        \hspace{-0.2cm}
      \includegraphics[width=.53\linewidth]{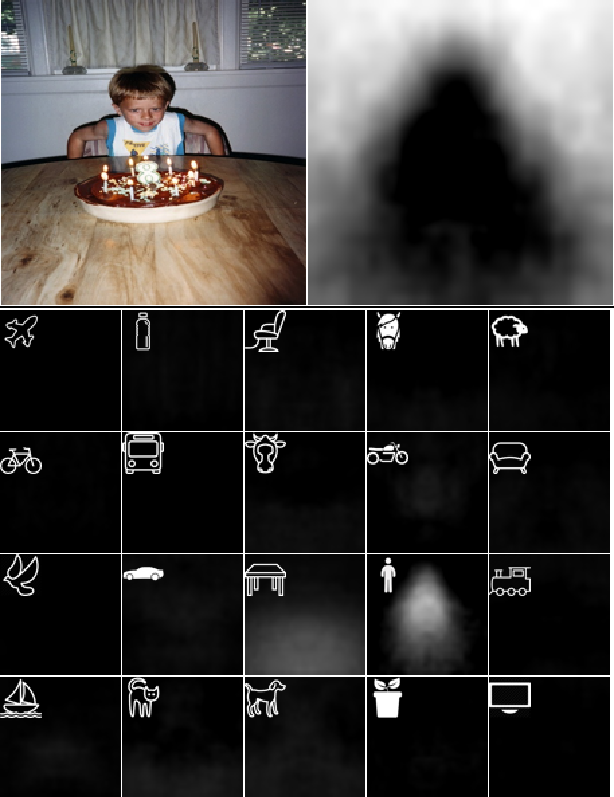}}
    &
    \hspace{-0.2cm}
    \includegraphics[width=.44\linewidth]{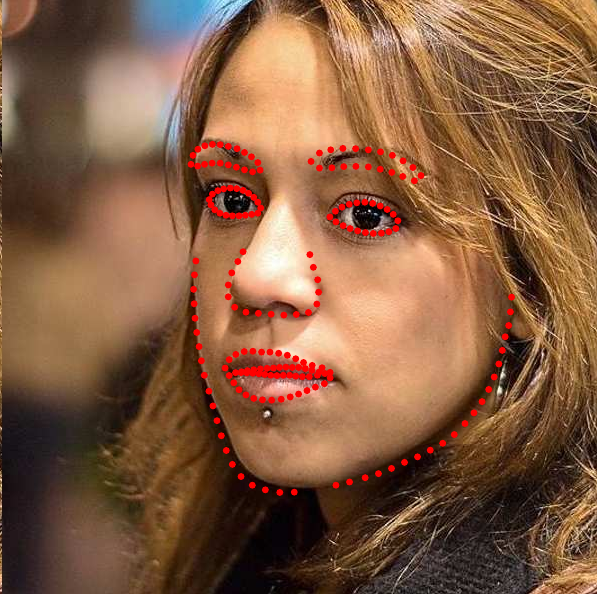}
    \vspace{0.12cm}
    \\
    & 
    \hspace{-0.3cm}
    \includegraphics[width=.44\linewidth]{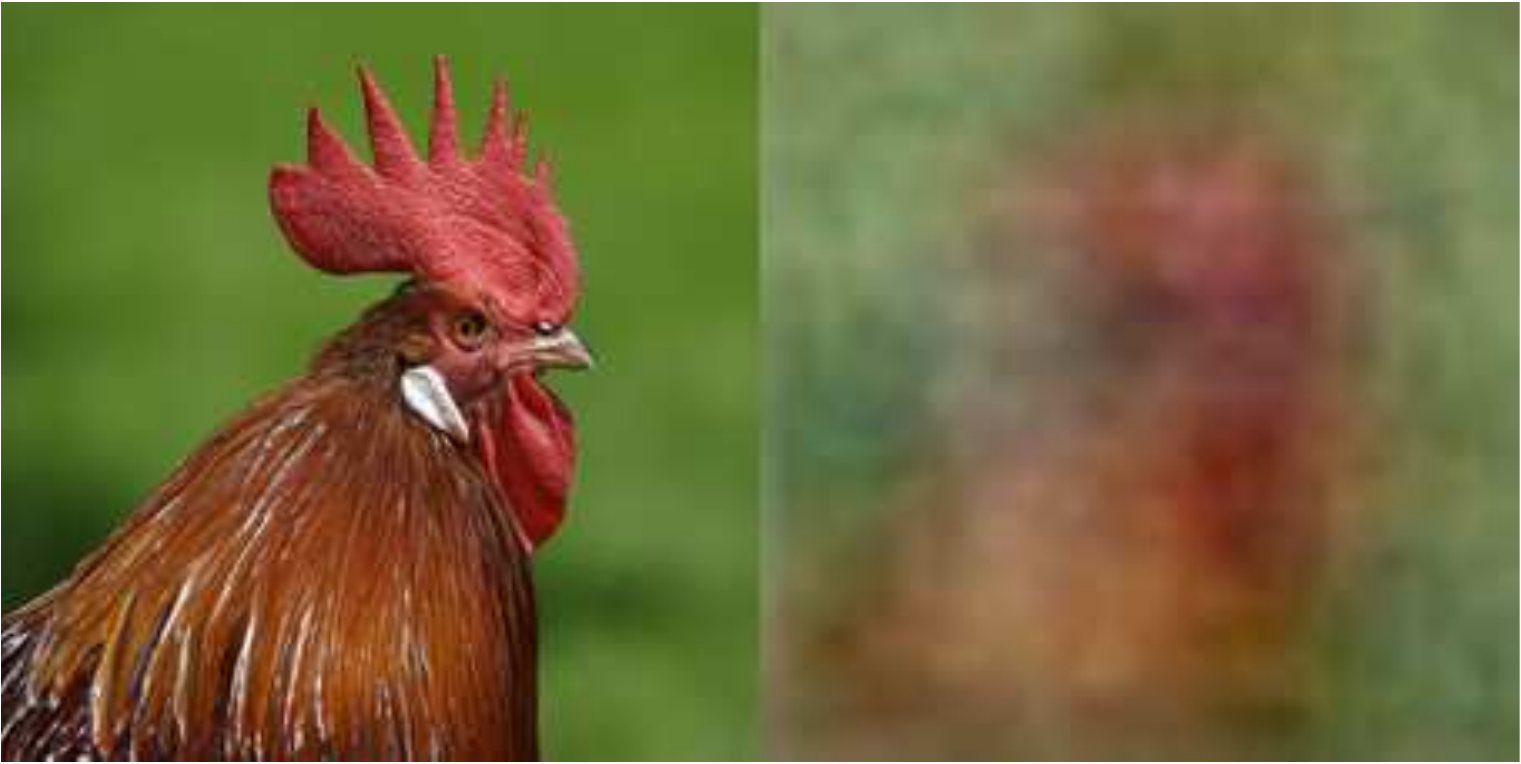}
    
  \end{tabular}
  \vspace{0.1cm}
  \caption{\small How many different local image properties can be predicted from a generic ConvNet representation using a \textit{linear} model? In this figure, given the ConvNet representation of an image, we have estimated three different local properties. Namely, semantic segmentation for background (top left) and 20 object classes (bottom left) present in PASCAL VOC dataset, 194 facial landmarks (top right) and RGB reconstruction of the original image (bottom right). One can see that a generic ConvNet representation optimized for ImageNet semantic classification has embedded high level of local information.}
  \label{fig:intro_teaser}  
\end{figure}

There is at least one alchemy associated with deep \CNNs. It occurs
when $\sim$100,000 iterations of SGD, in tandem with $\sim$1 million
labelled training images from ImageNet, transform the $\sim$60 million
randomly initialized weights of a deep \CNN{} into the best, by a huge
margin, performing known visual image classifier
\cite{Krizhevsky12,Girshick13,Oquab13,Sermanet13,Donahue14,Razavian14}.
Alongside this high-level alchemy is another related one w.r.t. the
image representations learnt by the fully connected layers of a deep
\CNN{} \cite{Girshick13,Oquab13,Donahue14,Razavian14}. These representations
are explicitly trained to retain information relevant to semantic
classes. But we show in this paper a striking fact, through various
tasks, that these \rep{}s also retain {\em spatial} information,
including the location of object parts and keypoints of object.


The notion of predicting spatial information using a \CNN{} itself is not new. 
Recent studies have introduced several approaches to extract spatial information
from an image with deep ConvNet. Some have trained a specialized
ConvNet to predict specific spatial information such as body parts and
facial landmarks \cite{Zhang14,Toshev14,EigenPF14}.
Others \cite{Sermanet13,BroxArxiv14,Long:arxiv:14} have shown that it
is possible to extract spatial correspondences using a generic \CNN{}
representation. But they consider representations from the
\CNN{} layers that only describe sub-patches of the whole
image. Then in a similar manner to a sliding window approach they have
an exhaustive spatial search, in tandem with their patch descriptor,
to find the locations.

Unlike those works, we show that a global image representation
extracted from the first fully connected layer of a \textit{generic}
ConvNet (\ie trained for predicting the semantic classes of ImageNet)
is capable of predicting spatial information \textit{without doing an explicit
search}.
In particular, we show that one can learn a linear regression function
(with results ranging from promising to good) 
from the representation to 
spatial properties: 2d facial landmarks, 2d object keypoints, RGB
values and class labels of individual pixels, see figure
\ref{fig:intro_teaser}.  We chose these experiments to highlight the
network's ability to reliably extract spatial information.

Why do we concentrate on the fully connected layers?  Prior work has
shown that these layers correspond to the most generic and compact
image representation and produce the best results, when combined with
a simple linear classifier, in a range of visual recognition tasks \cite{Azizpour14}.
Therefore the starting point of this work was to
examine what other information, besides visual semantic information,
is encoded and easily accessible from these representations.
%
%
 
%
The results we achieve for the tasks we tackle indicate that the
spatial information is implicitly encoded in the \CNN{} representation
we consider. Remember, the network has not been explicitly
encouraged to learn spatial information during the training.


The contributions of the paper are:
\begin{itemize}
\item For the first time we systematically demonstrate that spatial
  information is persistently transferred to the representation in the
  first fully connected layer of a generic ConvNet (section \ref{sec:experiments}). 
\item We show that one can learn a linear regression function from the
  \CNN{} representation to both object parts and local image
  properties. In particular, we demonstrate that it is possible to estimate 2d facial
  landmarks (section \ref{sec:facial_landmarks}), 2d object keypoints
  (section \ref{sec:object_keypoints}), RGB values (section
  \ref{sec:rgb_reconstruction}) and pixel level segmentations (section
  \ref{sec:sem_seg}). 
\item By using a simple \textit{look-back} method we achieved accurate
  predictions of facial landmarks on a par with state of the art (section \ref{sec:facial_landmarks}). 
\item We qualitatively show examples where semantically meaningful
  directions in the ConvNet representation space can be learned and
  exploited to accordingly alter the appearance of a face (section
  \ref{sec:semantic}).
\end{itemize}

Before describing our experiments and results in the next section we explain 
why spatial information can be ever retained 
and so easily accessed in the first fully-connected
layer of a generic \CNN.

%% file: Motivation.tex
\section{Flow of information through a \CNN{}}

A generic \CNN{} representation extracted from the first
fully-connected layer is explicitly trained to retain information
relevant to semantic class. The semantic classes in the training data are independent of spatial information and therefore this information, as it is deemed unnecessary to perform the task, should be removed or at least structured in such a manner that it does not conflict with the task.
%

The weights of a \CNN's convolutional layers encode a very large
number of compositional patterns of appearance that occur in the
training images. Thus, the multiple response maps output by a
convolutional layer indicate which appearance patterns occur in
different sub-patches (a.k.a. receptive fields) of a fixed size in the original image. The size
of these sub-patches increases as we progress through the
convolutional layers. When we come to the first fully-connected layer
the network must compress the set of response maps
(13$\times$13$\times$256 numbers assuming an \alexnet{} \CNN) produced
by the final convolutional layer into a mere 4096 numbers. The
compression performed seeks to optimize the ability of the network's
classification layer (with potentially some more intermediary
fully-connected layers) to produce semantic labels as defined by the
ImageNet classification task.

The weights of the first fully connected layer are, in general, not
particularly sparse. Therefore the \textit{what} and \textit{where}
explicitly encoded in the convolutional layers are aggregated, merged
and conflated into the output nodes of this first fully connected
layer. At this stage it is impossible to backtrack from these
responses to spatial locations in the image. Nevertheless, we show it
is possible to predict from this global image descriptor, using linear
regressors, the spatial locations of object parts and keypoints and
also pixel level descriptors such as colour and semantic class.

%% file: Experiment.tex
\section{Experiment}\label{sec:analysis}
\label{sec:experiments}

We study two families of tasks to explore which spatial information
resides in the ConvNet representation:
\begin{itemize}
\item Estimate the $(x,y)$ coordinate of an item in an image.
\item Estimate the local property of an image at $(x,y)$.
\end{itemize}

Given the \CNN{} representation of an image for the first
task, we
\begin{enumerate*}[label=\itshape\roman*\upshape)]
  \item estimate the coordinates of facial landmarks in three
    challenging datasets \cite{LeBLBH12,BelhumeurJKK11,SagonasTZP13},
    and
\item predict the positions of object keypoints. We use the
  annotations \cite{Bourdev:iccv:2009} from the Pascal VOC 2011
  dataset as our testbed.
\end{enumerate*}
While for the second task, given a \CNN{} representation, we
\begin{enumerate*}[label=\itshape\roman*\upshape)]
  \item predict the RGB values of every pixel in the original image
    (we use the ImageNet validation set as our test set), and
  \item predict the semantic segmentation of each pixel in the
    original image (VOC 2012 Pascal dataset).
\end{enumerate*}


\subsection{Experimental setup}
In all our experiments we use the same \CNN. It has the \alexnet{}
architecture \cite{Krizhevsky12} and is trained on
ImageNet \cite{ilsvrc-2013} using the reference implementation provided
by \caffe ~\cite{Jia14}. Our image representation then corresponds to
the responses of the first fully connected layer of this network
because of its compactness and ability to solve a wide range of
recognition tasks \cite{Girshick13,Oquab13,Donahue14,Razavian14,Azizpour14}. We
will denote this representation by $\bff$. Then the only
post-processing we perform on $\bff$ is to $l_2$ normalize it.

For every scalar quantity $y$ we predict from $\bff$, we do so
with a linear regression model:
\begin{align}
  y \approx \bw^T \bff + w_0
\end{align}
We use a ridge regularised linear model because of its simplicity and
for the following reason. All the class, pose
and semantic information does exist in the original RGB image, as the
human vision proves, but it is not easily accessible and especially
not through linear models. However, we want to study if all this
information is still encoded in the \CNN{} representation, but in a
much more accessible way and this is demonstrated by the use of a
linear model compared to a much more capable prediction algorithm.

\begin{table}[t]
  \setlength{\tabcolsep}{3pt}
  \centering
\begin{tabular}{@{}l c c c@{}}
\toprule
  & Helen \cite{LeBLBH12} & LFPW \cite{BelhumeurJKK11} & IBUG \cite{SagonasTZP13}\\
\midrule
Dataset Bias & 0.501 & 0.242 & 0.352\\
\midrule
RGB + ridge  & 0.096 & 0.074 & 0.160 \\
STASM \cite{MilborrowN08} & 0.111 & - & - \\
CompASM \cite{LeBLBH12} & 0.091 & - & - \\
\midrule
\CNN{} + ridge & 0.065 & 0.056 & 0.096 \\
\midrule
RCPR \cite{Burgos-ArtizzuPD13} & 0.065  & 0.035 & - \\
SDM  \cite{XiongT13} & 0.059 & 0.035 & 0.075 \\
ESR  \cite{CaoWWS12} & 0.059 & 0.034 & 0.075 \\
ETR  \cite{KazemiS14}& 0.049 & 0.038 & 0.064 \\
\midrule
\CNN{} + \textit{look-back} & 0.058 & 0.049 & 0.074 \\
\bottomrule
\end{tabular}
\vspace{.3cm}
\caption[]{\small Evaluation of facial landmark estimation on three
  standard face datasets and comparison with baslines and recent state
  of the art methods. The error measure is the average distance
  between the predicted location of a landmark and its ground truth
  location. Each error distance is normalized by the
  inter-occular distance.}
\label{tab:face_landmarks_result}
\end{table}

There are, of course, numerous ways we can estimate the coefficients $(\bw, w_0)$ from
labelled training. Assume
that we have labelled training data $(y_1, \bff_1), \ldots (y_n,
\bff_n)$ where each $y_i \in \mathbb{R}$ and $\bff \in
\mathbb{R}^d$ ($d=4096$). The optimal values for $(\bw, w_0)$ are then found
solving this optimization problem
\begin{align}
  (\bw^*, w^*_0) = \underset{\bw, w_0}{\arg\min} \sum_{i=1}^n (y_i - \bw^T \bff_i - w_0) +
  \lambda \|\bw\|^2
\end{align}
The closed form solution to this optimization problem is easily shown
to be:
\begin{align}
  \bw^* = \left(X^T X + \lambda I \right)^{-1} X^T \by
  \quad \text{and} \quad w_0^* = \frac{1}{n} \sum_{i=1}^n y_i
  \label{eq:ridge}
\end{align}
where
\begin{align}
  X = \begin{pmatrix}
    \leftarrow \bff_1^T \rightarrow\\
    \leftarrow \bff_2^T \rightarrow\\
   \vdots\\
   \leftarrow \bff_n^T \rightarrow\\
  \end{pmatrix}\qquad\text{and}\qquad
  \by = \begin{pmatrix}
    y_1\\
    y_2\\
    \vdots\\
    y_n
    \end{pmatrix}
\end{align}
and it is assumed the columns of $X$ have been centred. In all our
experiments we set the regularization parameter $\lambda$ with
four-fold cross-validation.

\begin{figure}[t!]
  \centering
  \includegraphics[width=0.5\textwidth]{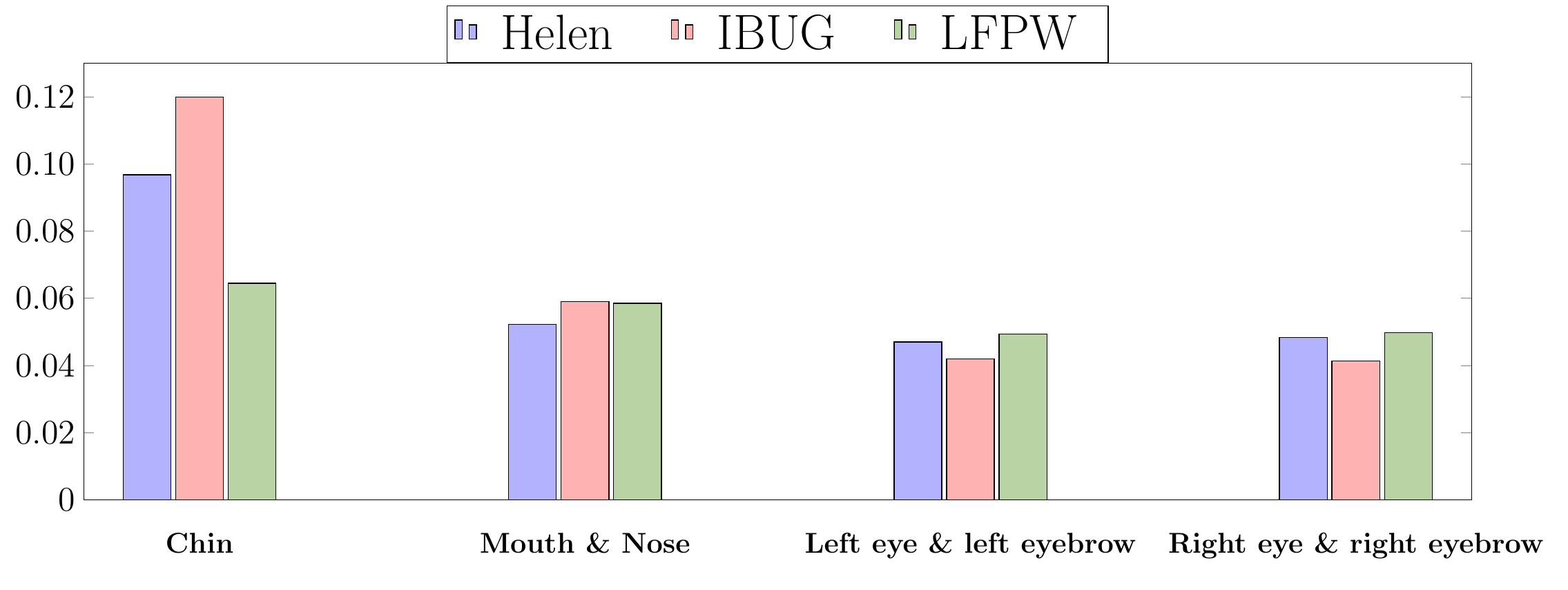}
  \caption{\small The normalized prediction error for different
    subsets of the landmarks after \textit{look-back}
    (see the caption of table \ref{tab:face_landmarks_result} for the
    error measure).
    The error is shown for three different face datasets. 
    Since the bounding box
    around the chin is bigger than the bounding box around the other
    parts, the error for chin is higher than the rest of facial
    landmarks.}
  \label{fig:landmark_error}
\end{figure}

 \def\pwid{.487\linewidth}
 \begin{figure*}
   \centering
   \begin{tabular}{@{}cc@{}}
     \includegraphics[width=\pwid]{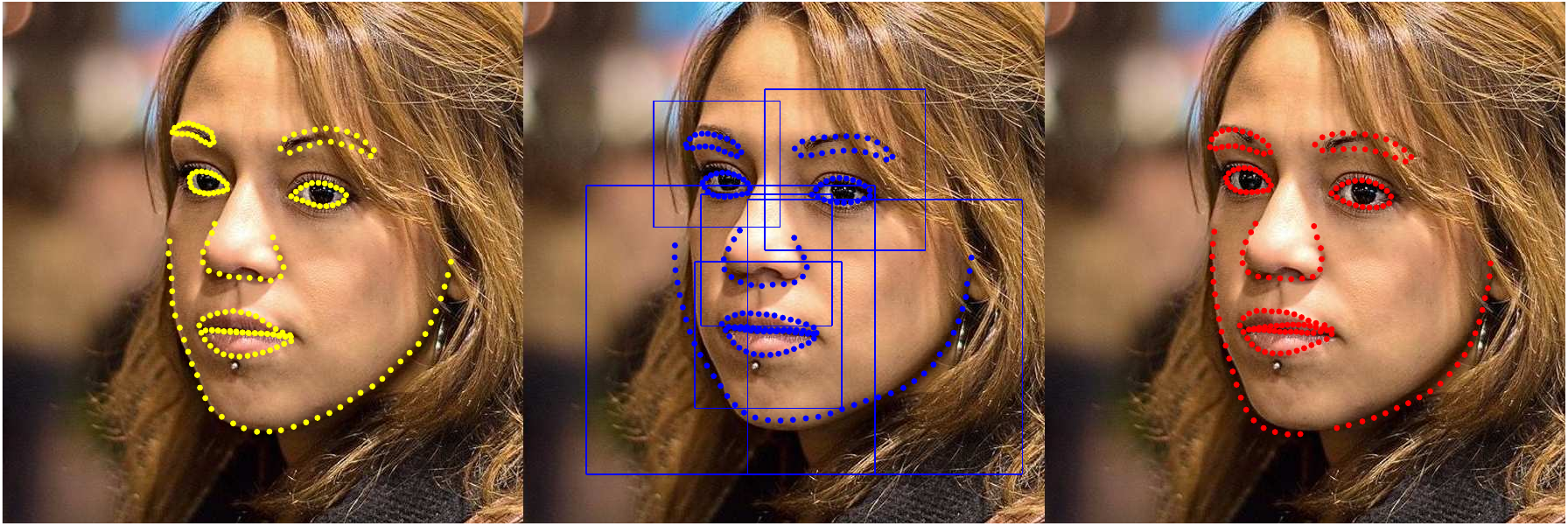} &
     \includegraphics[width=\pwid]{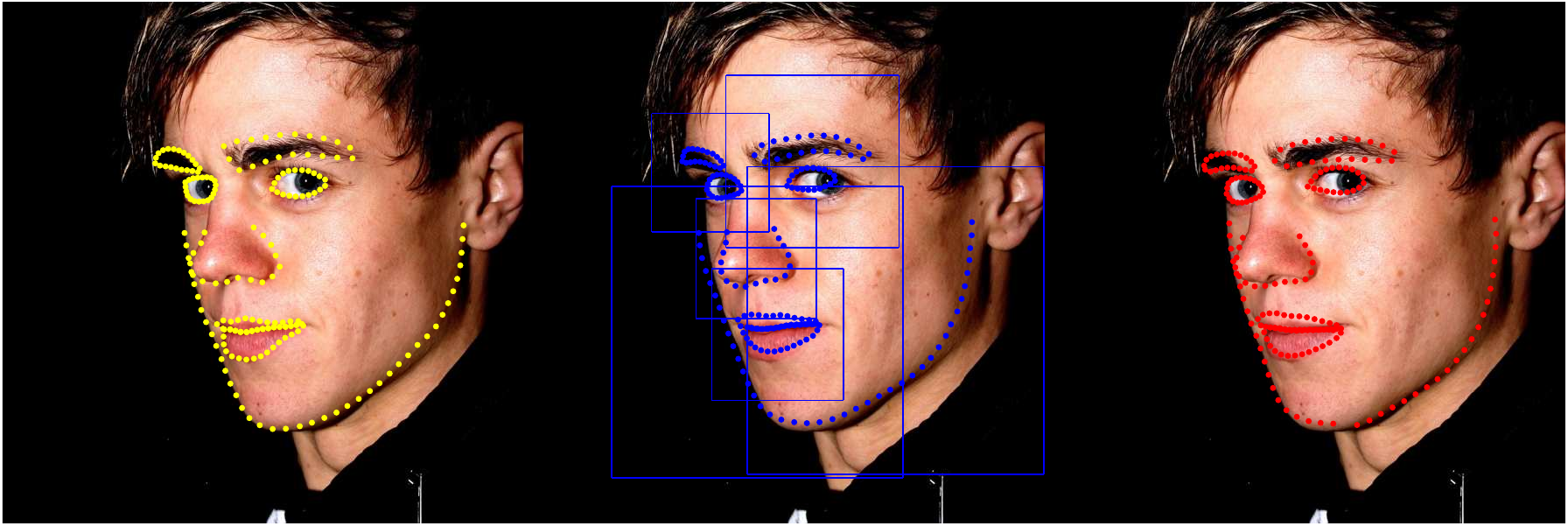}\\
     \includegraphics[width=\pwid]{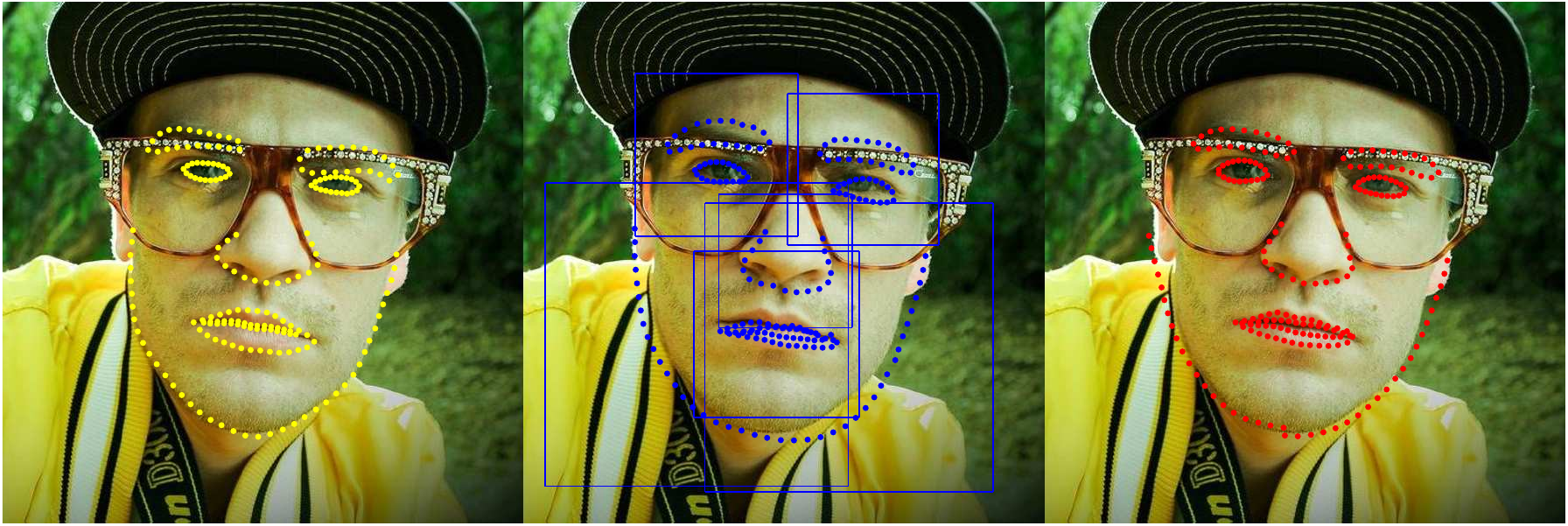} &
     \includegraphics[width=\pwid]{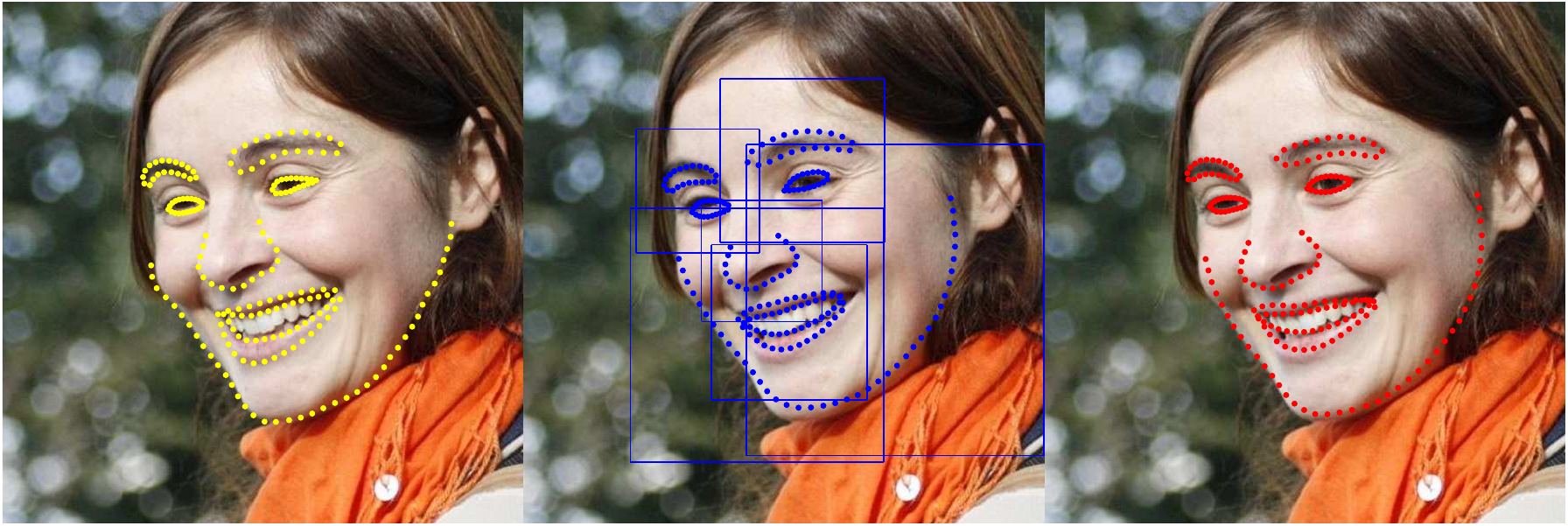}\\
     \includegraphics[width=\pwid]{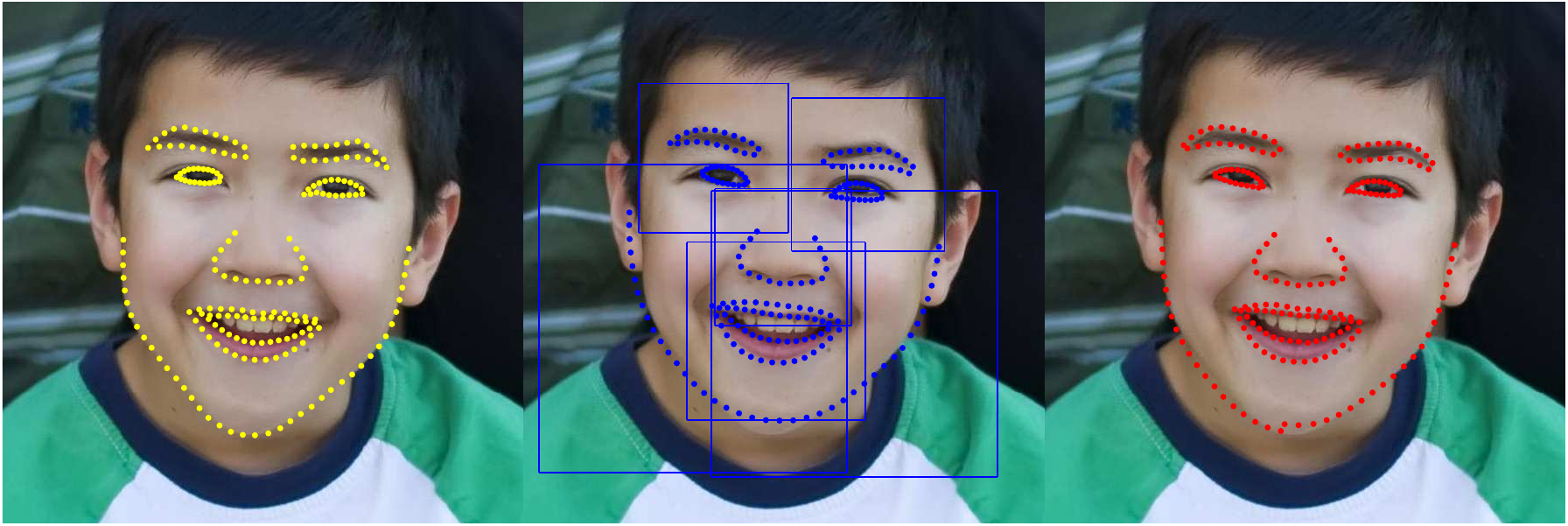} &
     \includegraphics[width=\pwid]{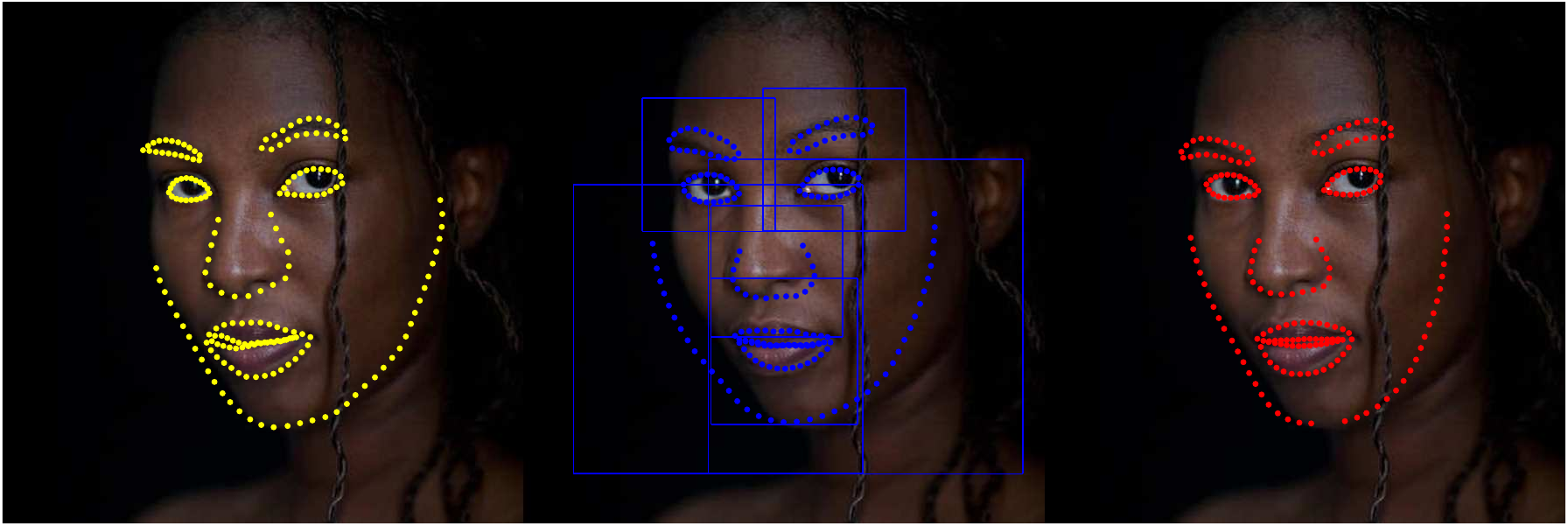}\\
     \includegraphics[width=\pwid]{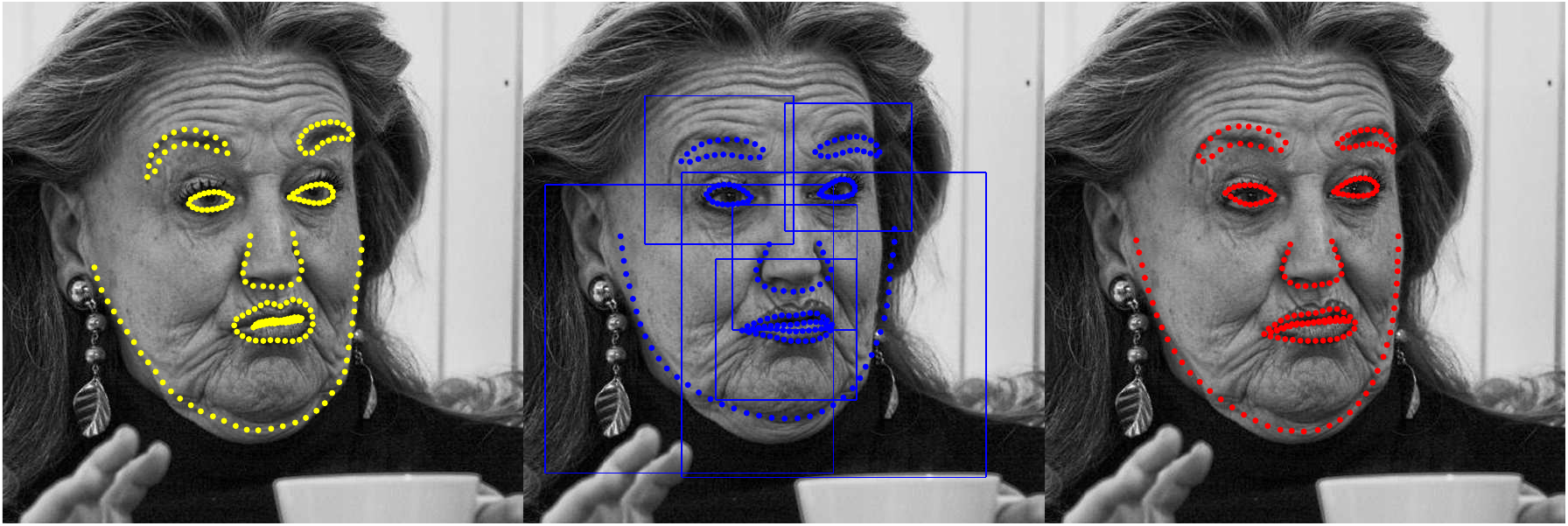} &
     \includegraphics[width=\pwid]{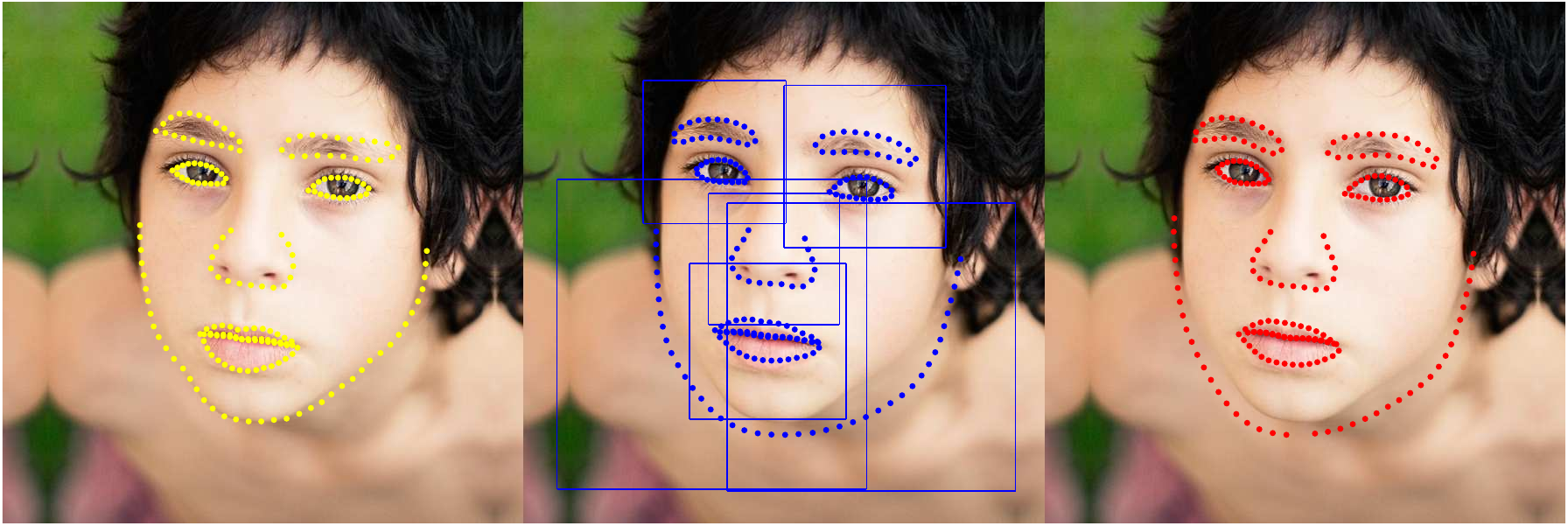}
   \end{tabular}
 \caption{\small The landmarks predicted by our linear regressors for
   eight different images from the Helen Dataset. The leftmost image
   in each triplet shows the ground truth. The middle image shows the
   landmarks predicted by the linear regression functions from the
   \CNN{} representation of the whole image to landmark
   coordinates. In the middle image the bounding boxes, defined by the
   initial predictions for the landmarks, used by \textit{look-back}
   method are also shown.  The rightmost image shows the final
   landmark predictions made by the \textit{look-back} method.}
 \label{fig:landmark_prediction}
 \end{figure*}

\subsubsection{Facial Landmarks}
\label{sec:facial_landmarks}

The first problem we address is the popular task of 2d facial landmark
detection. Facial landmark detection is interesting since a large body
of work has been applied to it. To train our landmark estimation
model, for each landmark we estimate two separate linear regression
functions, one for the $x$-coordinate of the landmark and one for the
$y$-coordinate. Therefore we estimate the $(x,y)$ coordinates of all
the $L$ landmarks from the image's \CNN{} representation, $\bff$,
with

\begin{align}
  \hat{\bx} = W_{\text{\tiny landmarks}} \bff + \bw_{\text{\tiny
      landmarks},0}
  \label{eq:facial_landmarks}
\end{align}
where $W_{\text{\tiny landmarks}}\in \mathbb{R}^{2L \times d}$ and
$\bw_{\text{\tiny landmarks},0}\in \mathbb{R}^{2L}$. Remember each row
of $W_{\text{\tiny landmarks}}$ is learnt independently via the ridge
regression solution of equation (\ref{eq:ridge}). For the rest of the
tasks explored in this section we use a similar formulation to the one
just described so we will not introduce new notation to describe them.

Table \ref{tab:face_landmarks_result} details the average errors, of
our approaches and other methods, in the predicted location of the
landmarks on three standard datasets: Helen \cite{LeBLBH12}, LFPW
\cite{BelhumeurJKK11} and IBUG \cite{SagonasTZP13}. The reported
errors are normalized by the distance between two eyes in the image
according to the standard practice in the field \cite{KazemiS14}. The
table reports the performance of both the baseline predictors of
linear ridge regression from RGB and a random predictor and recent
high performing systems
\cite{Burgos-ArtizzuPD13,XiongT13,KazemiS14,CaoWWS12} which generally
involve learning a complicated non-linear function from RGB to the
landmarks. Our predictor, \CNN+ridge, produces a significantly better
estimate than the baselines and its performance is comparable with state-of-the-art methods specifically designed to solve this problem. Our result indicates
that the locations of landmarks can be reliablly extracted from the
\CNN{} representation.

\begin{table*}[th!]
  \scriptsize
  \centering
  \setlength{\tabcolsep}{2pt}
\begin{tabular}{@{}l *{20}{c@{\hspace*{5pt}}}  r}
\toprule
  & airplane & bike & bird & boat & bottle & bus & car & cat & chair & cow & table & dog & horse & motorbike & person & plant & sheep & sofa & train & tv & \textbf{mean} \\
\midrule
SIFT & 17.9 & 16.5 & 15.3 & 15.6 & 25.7 & 21.7 & 22.0 & 12.6 & 11.3 & 7.6 & 6.5 & 12.5 & 18.3& 15.1& 15.9& 21.3& 14.7& 15.1& 9.2& 19.9& 15.7\\
SIFT+prior & 33.5 & 36.9 & 22.7 & 23.1 & 44.0 & 42.6& 39.3& 22.1& 18.5& 23.5& 11.2& 20.6& 32.2& 33.9& 26.7& 30.6& 25.7& 26.5& 21.9& 32.4& 28.4\\
\midrule
\CNN{} + ridge & 21.3 &  25.1 &  22.7 &  16.4 &  47.3 &  27.2 &  29.9 &  25.4 &  19.7 &  26.3 &  22.0
&  27.1 &  25.5 &  21.8 &  33.8 &  41.0 &  28.2 &  23.0 &  23.9 & 47.3 & 27.8\\
\midrule
Conv5 + sliding window \cite{Long:arxiv:14} & 38.5 & 37.6 & 29.6 & 25.3 & 54.5 & 52.1 & 28.6 & 31.5 & 8.9 & 30.5 & 24.1 & 23.7 & 35.8 & 29.9 & 39.3 & 38.2 & 30.5 & 24.5 & 41.5 & 42.0 & 33.3\\
\bottomrule
\end{tabular}
\vspace{0.1cm}
\caption{\small Quantitative evaluation of our keypoint estimation for
  general objects on VOC11. The performance measure is the average PCK
  score with $\alpha=0.1$.}
  \vspace{-5mm}
 \label{tab:VOC11_keypoint_result}
\end{table*}

\CNN{}+ridge inherently loses around $\pm$10 pixel accuracy due to
the pooling and strides in the first and second convolutional layers
of the ConvNet.  However, we can overcome this limitation in a simple
manner which we term the \textit{look-back trick}. We partition the
landmarks into different subsets (such as chin, left eye and left
eyebrow, right eye and right eyebrow, and then mouth and nose), figure
\ref{fig:landmark_prediction} shows some examples. We let the
predicted position of each set of landmarks, using equation
\ref{eq:facial_landmarks}, define a square bounding box containing
them with some margin based on the maximum prediction error for the
landmarks in the training set. We then extract the \CNN{}
representation for this sub-image and use linear regression, as
before, to estimate the coordinates of the landmarks in the
bounding-box. This simple trick significantly boosts the accuracy of
the predictions, and allows us to outperform all the s.o.a. methods
except for the recent work of \cite{KazemiS14}. The more sets we have
in the partition the better results we get. We used six sets for each
dataset. See figure \ref{fig:landmark_prediction} for qualitative
examples of the result of this method on sample images from the Helen
dataset.  Figure \ref{fig:landmark_error} also shows the prediction
errors for different parts after look-up for three face datasets.

\subsubsection{Object Keypoints}
\label{sec:object_keypoints}
In our next task we predict the location of object keypoints. These
keypoints exhibit more variation in their spatial location than facial
landmarks. We use the keypoint annotations provided by
\cite{Bourdev:iccv:2009} for 20 classes of PASCAL VOC 2011.  We make
our predictions using exactly the same basic approach as for facial
landmarks. The classes of PASCAL task include many deformable objects
(dog, cat, human, \etc) and objects which have high intra-class
variation (bottle, plant, \etc) which makes the problem of key point
detection extremely difficult. In order to model these keypoints a
separate set of regressors is learnt for each object.

\begin{figure}[tb]
\centering
\includegraphics[width=\linewidth]{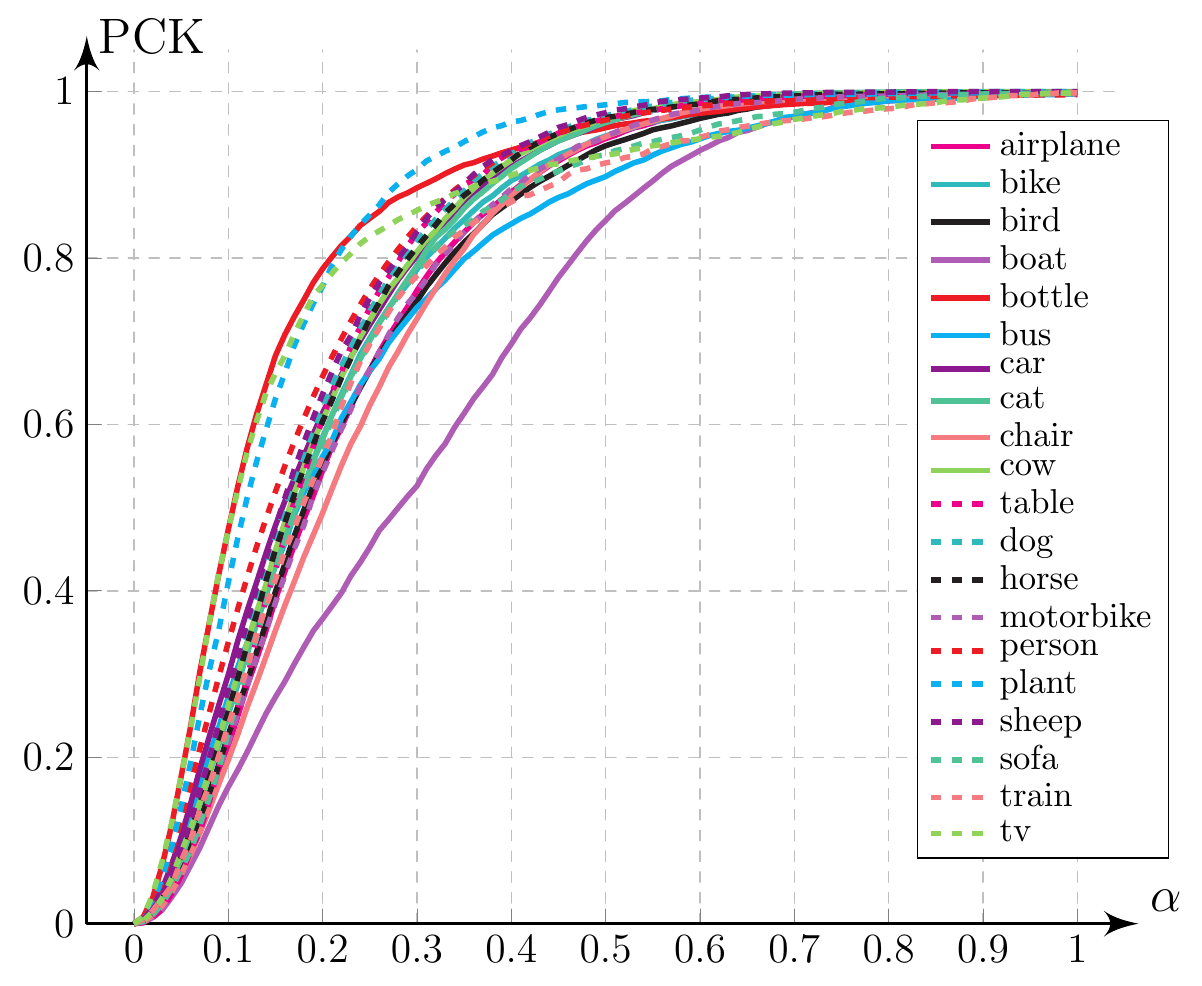}
    \caption{PCK evaluation for keypoint prediction of 20 classes of PASCAL VOC 2011.}
 \label{fig:keypoint_PCK_alpha}
\end{figure}

\def\pwid{.18\linewidth}
\begin{figure*}[htpb]
  \centering
    \centering
  \begin{tabular}{@{}ccccc@{}}
    \includegraphics[width=\pwid]{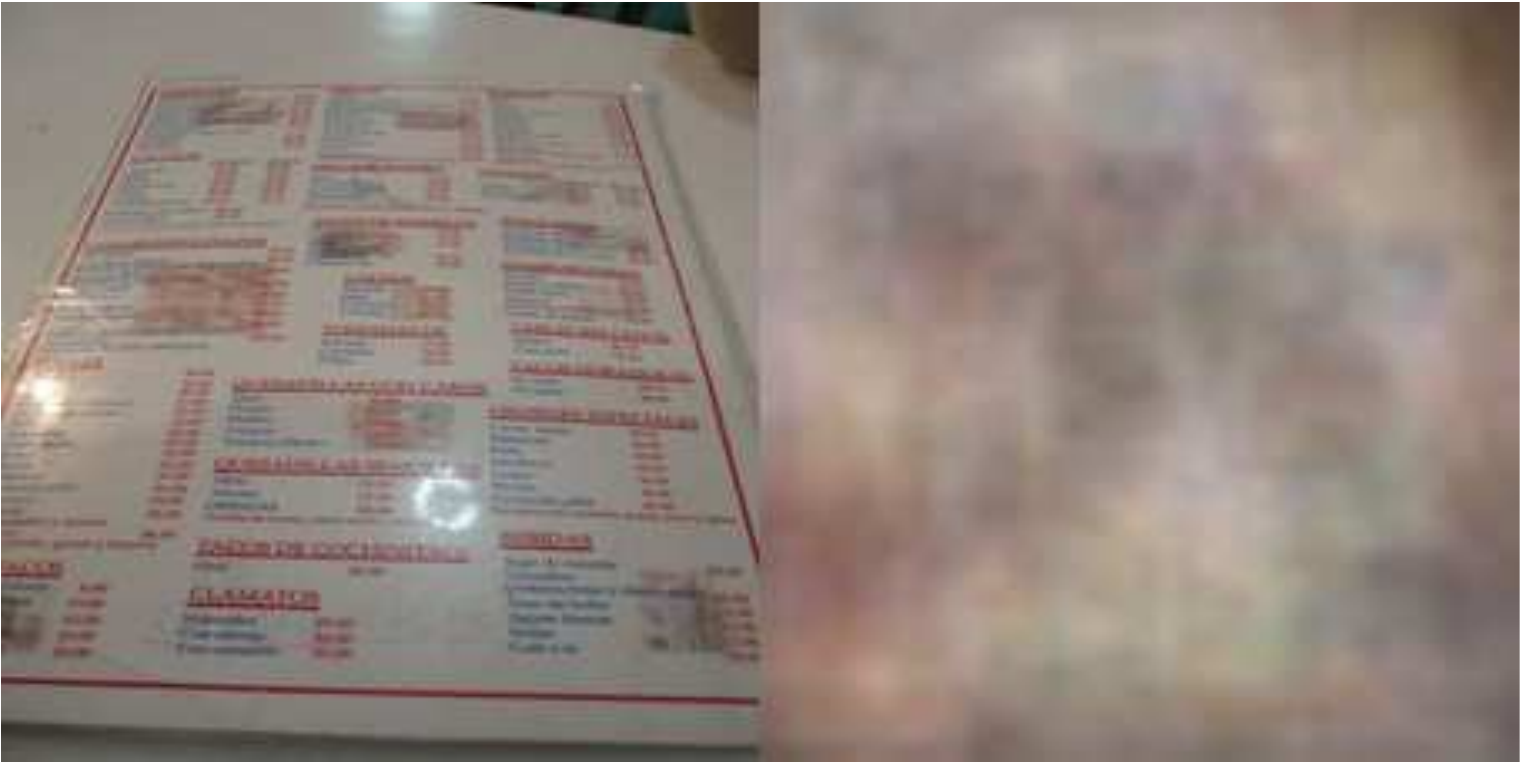}
    &
    \includegraphics[width=\pwid]{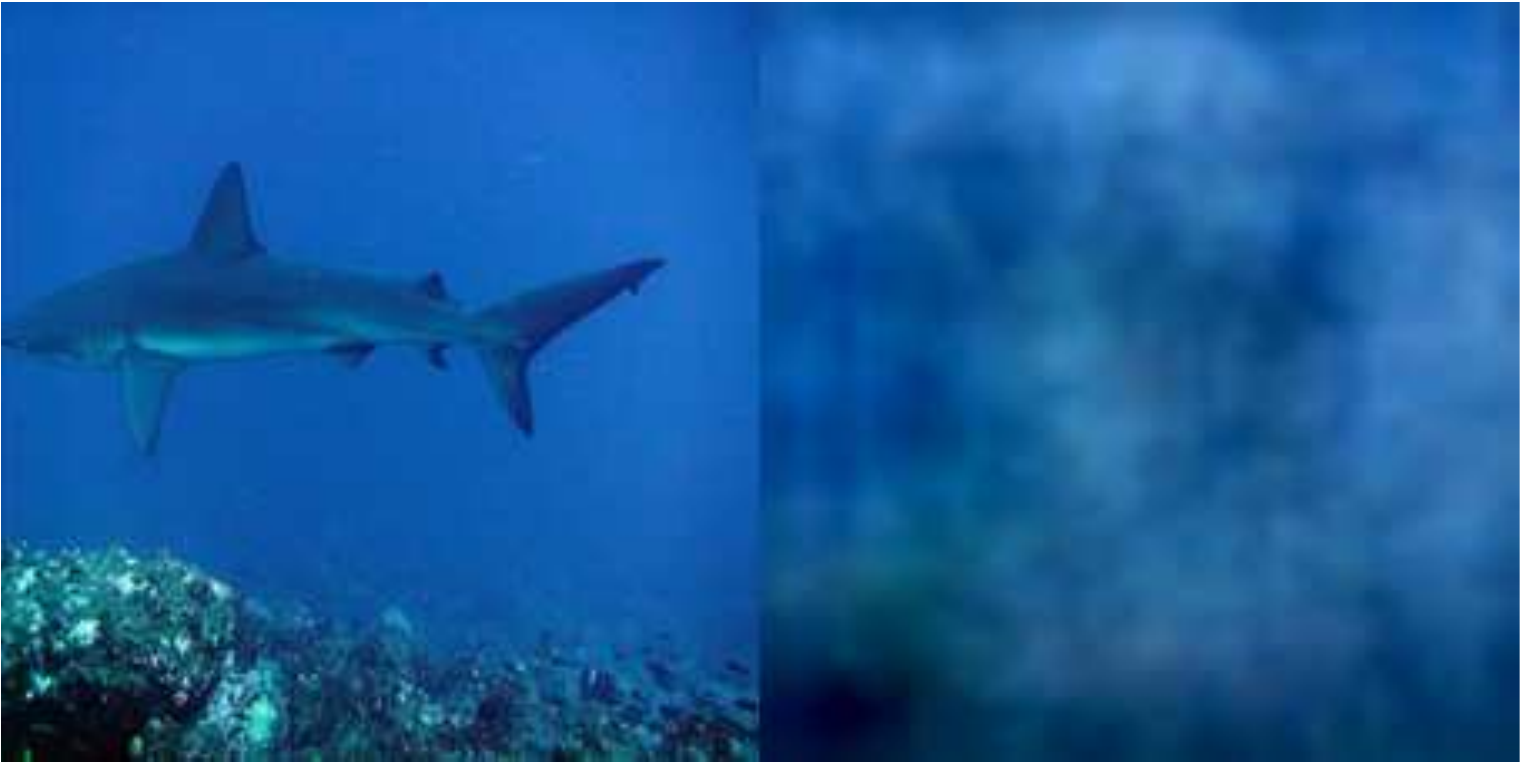}
    &
    \includegraphics[width=\pwid]{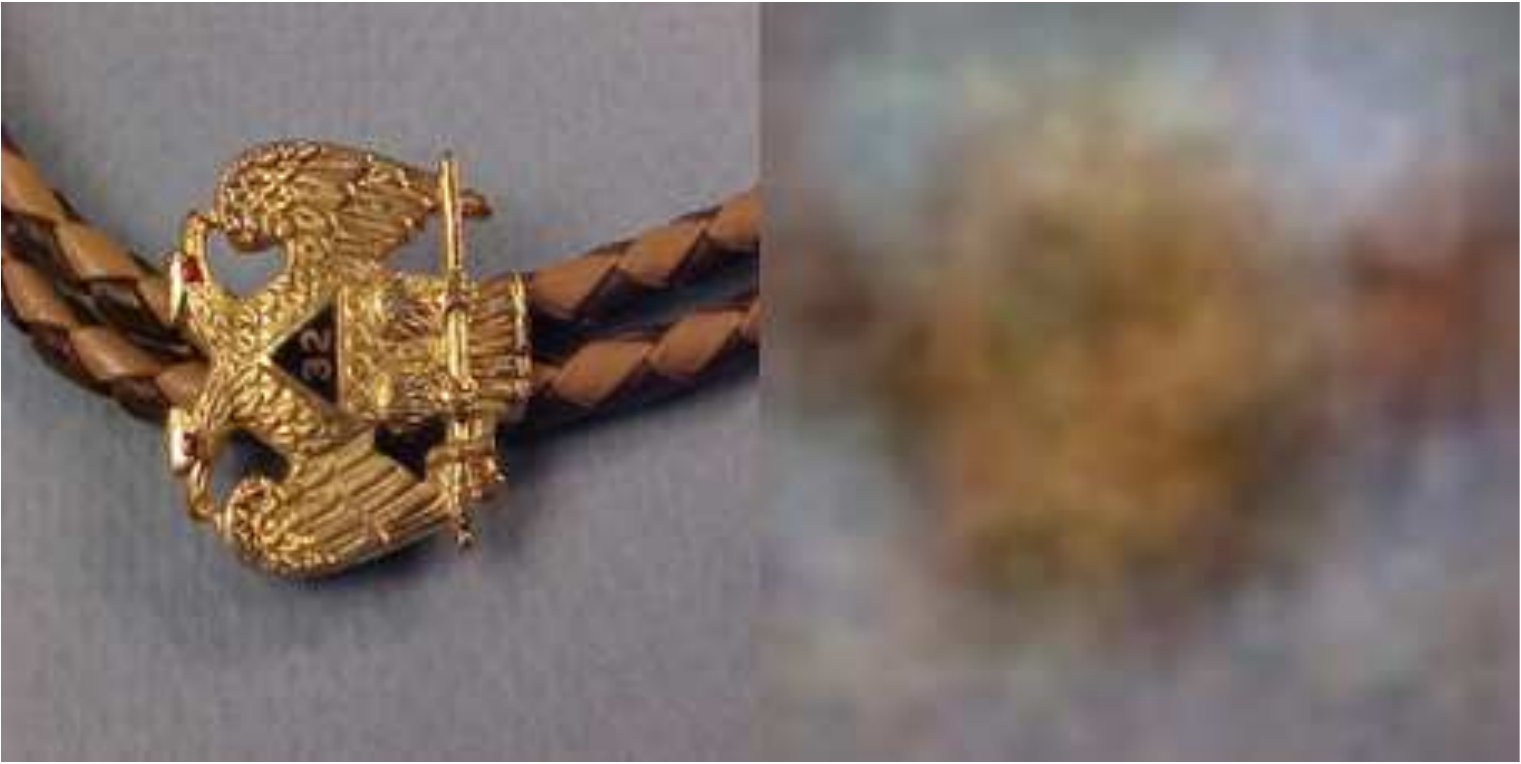}
    &
    \includegraphics[width=\pwid]{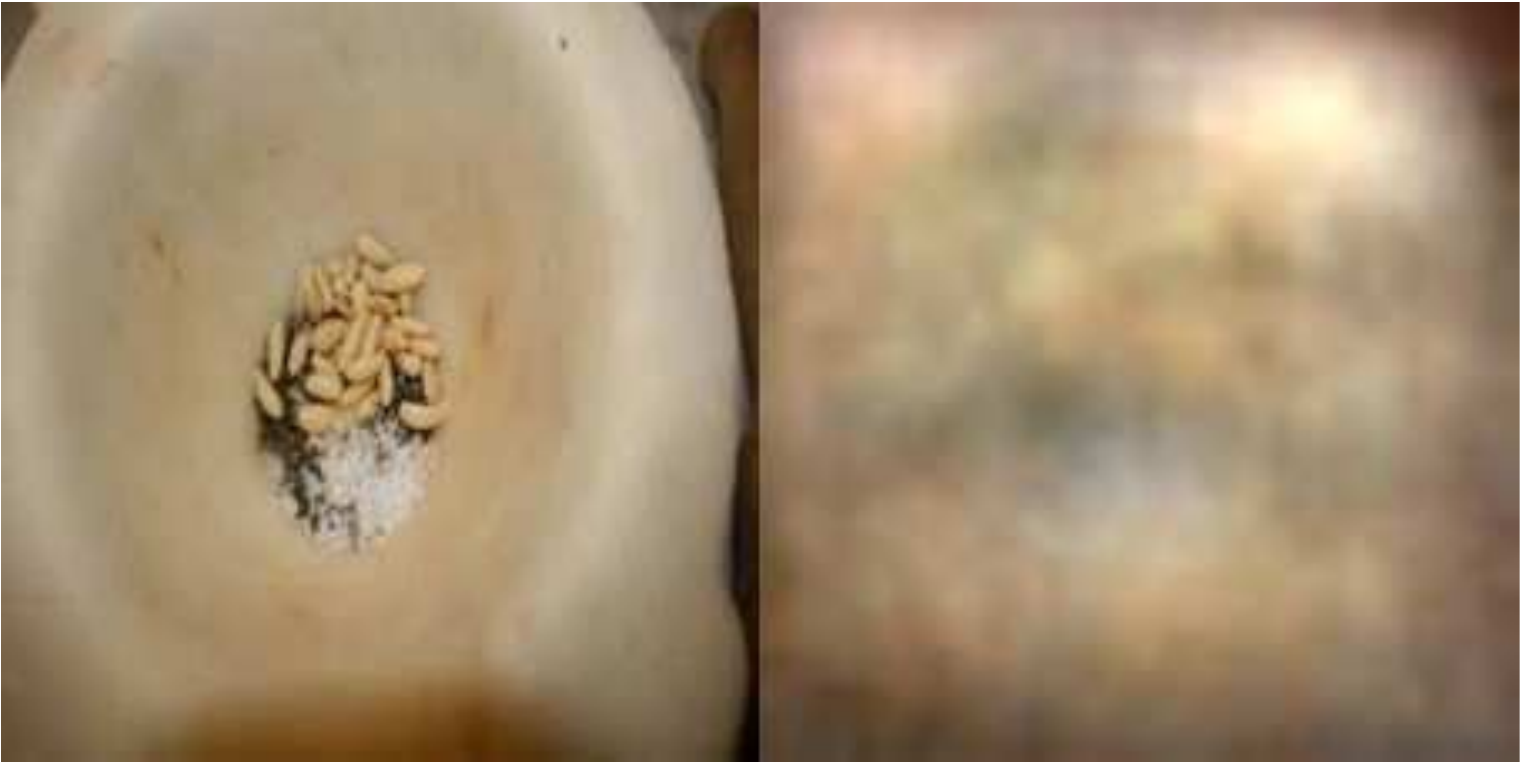}
    &
    \includegraphics[width=\pwid]{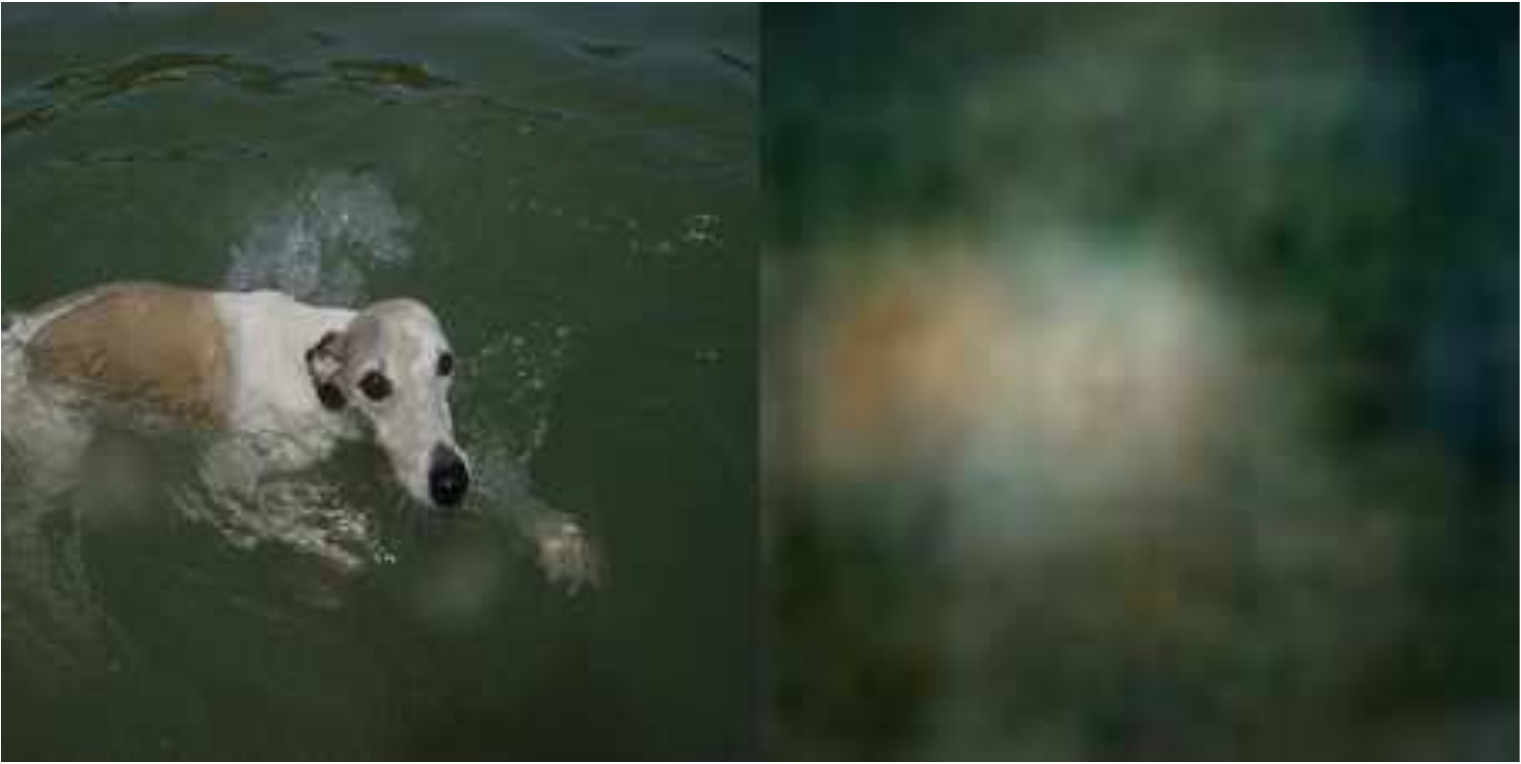}\\
    \includegraphics[width=\pwid]{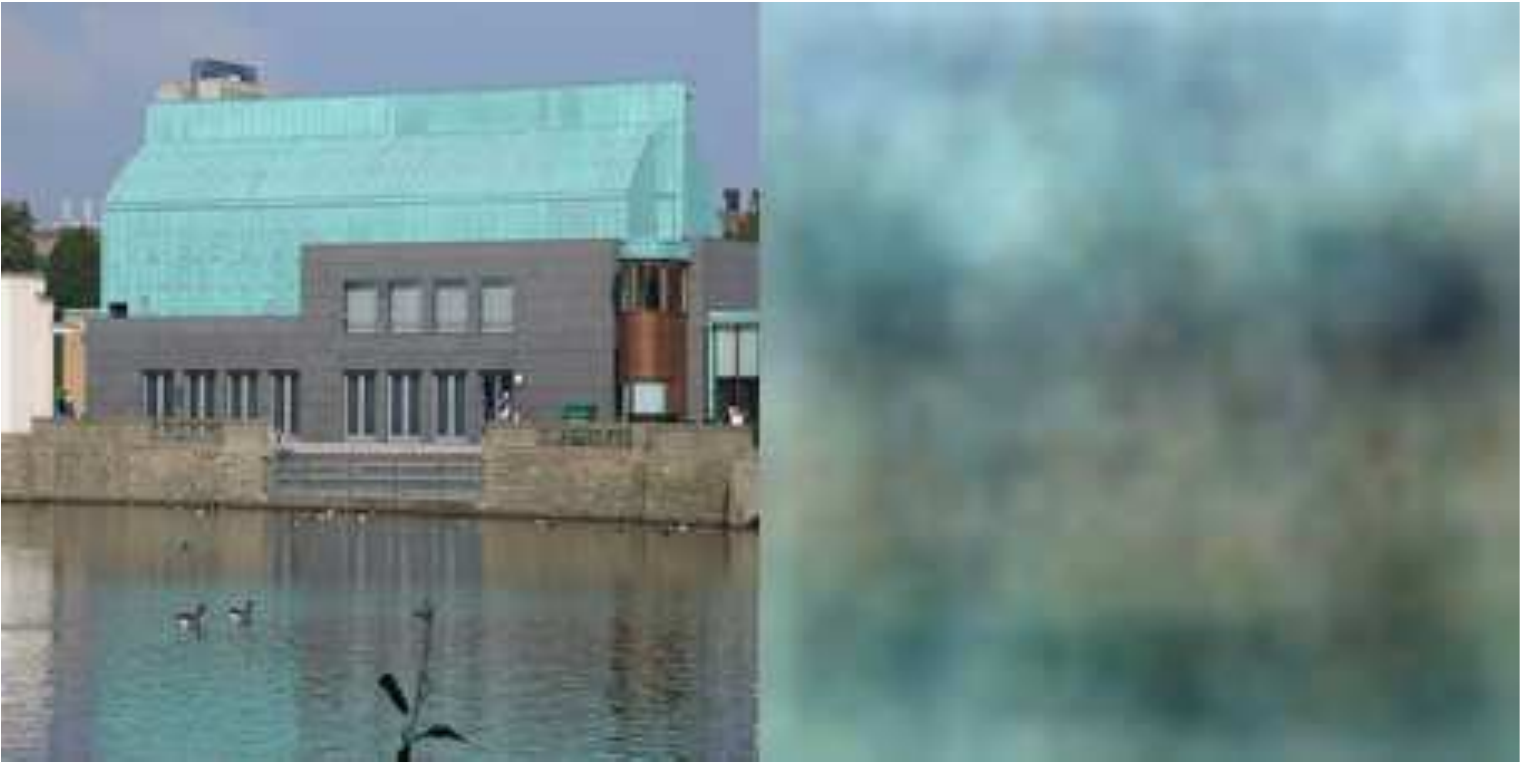}
    &
    \includegraphics[width=\pwid]{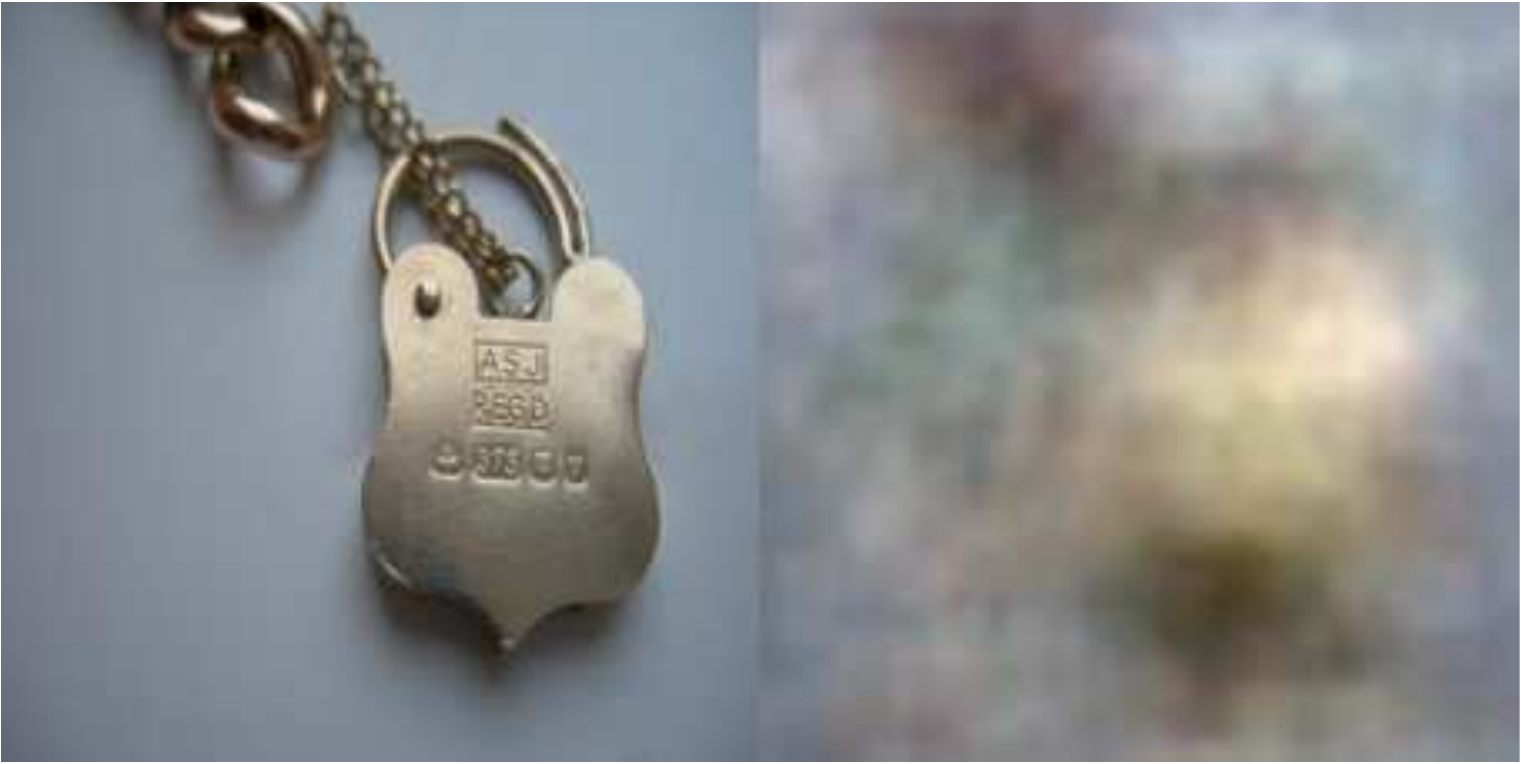}
    &
    \includegraphics[width=\pwid]{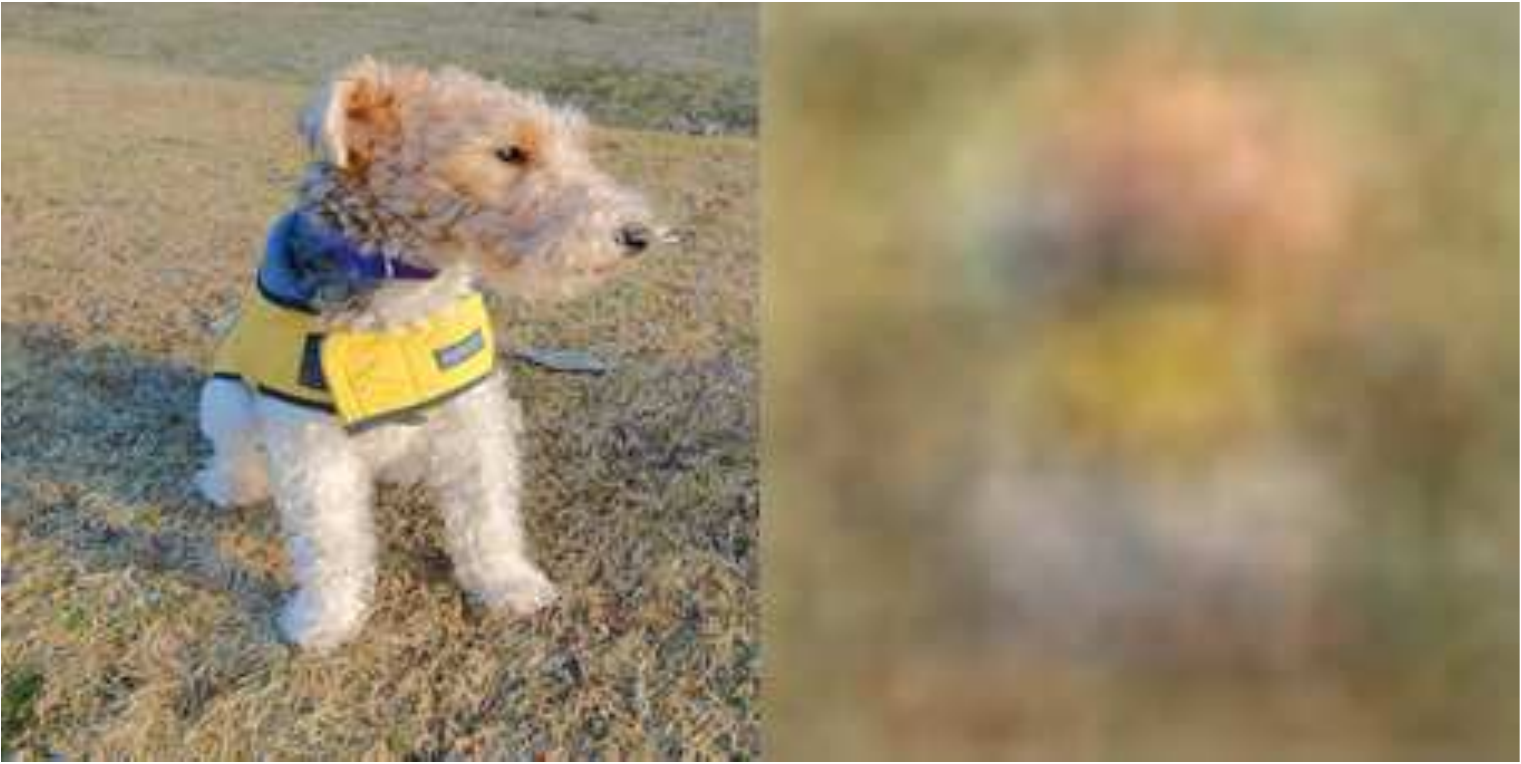}
    &
    \includegraphics[width=\pwid]{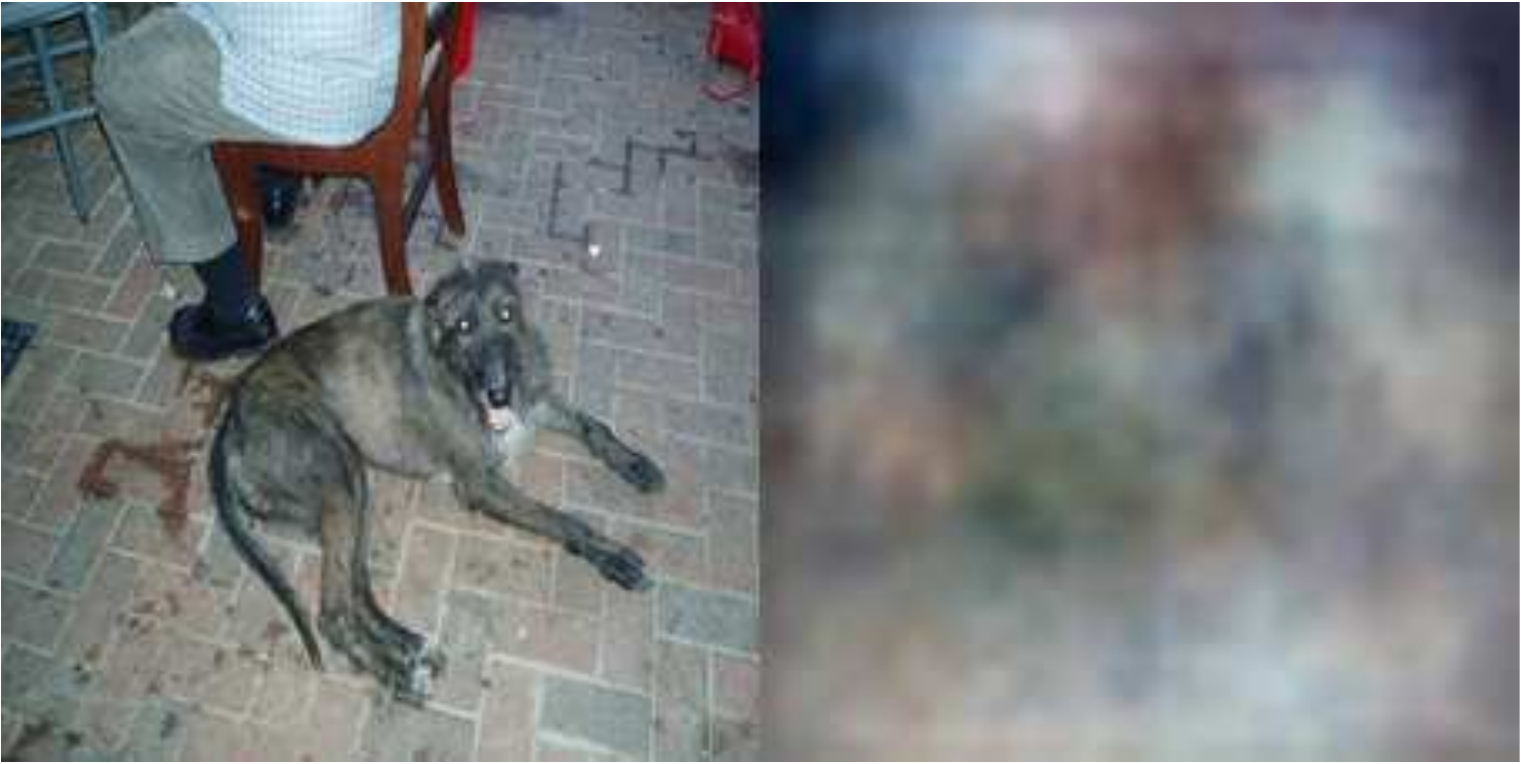}
    &
    \includegraphics[width=\pwid]{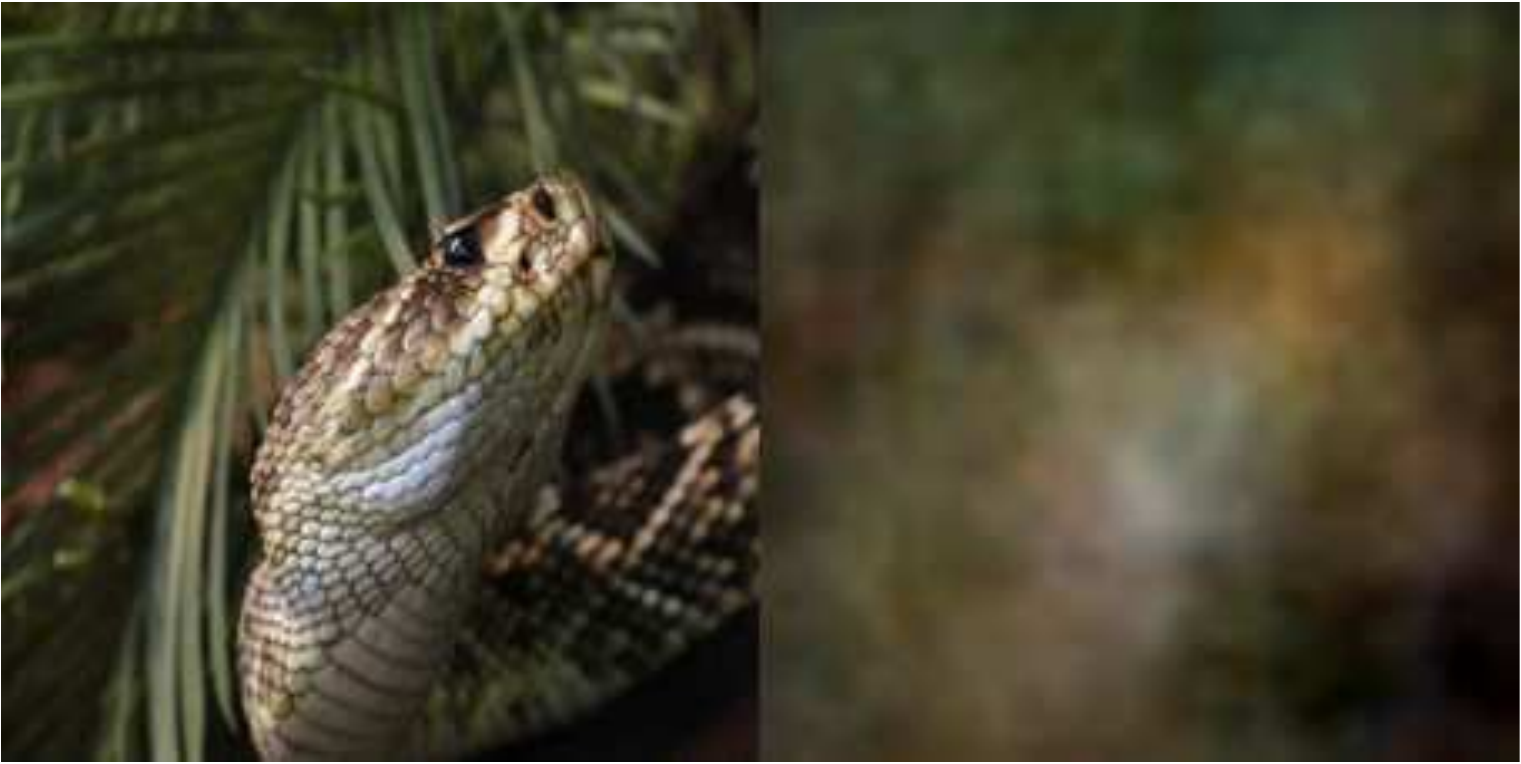}\\
        \includegraphics[width=\pwid]{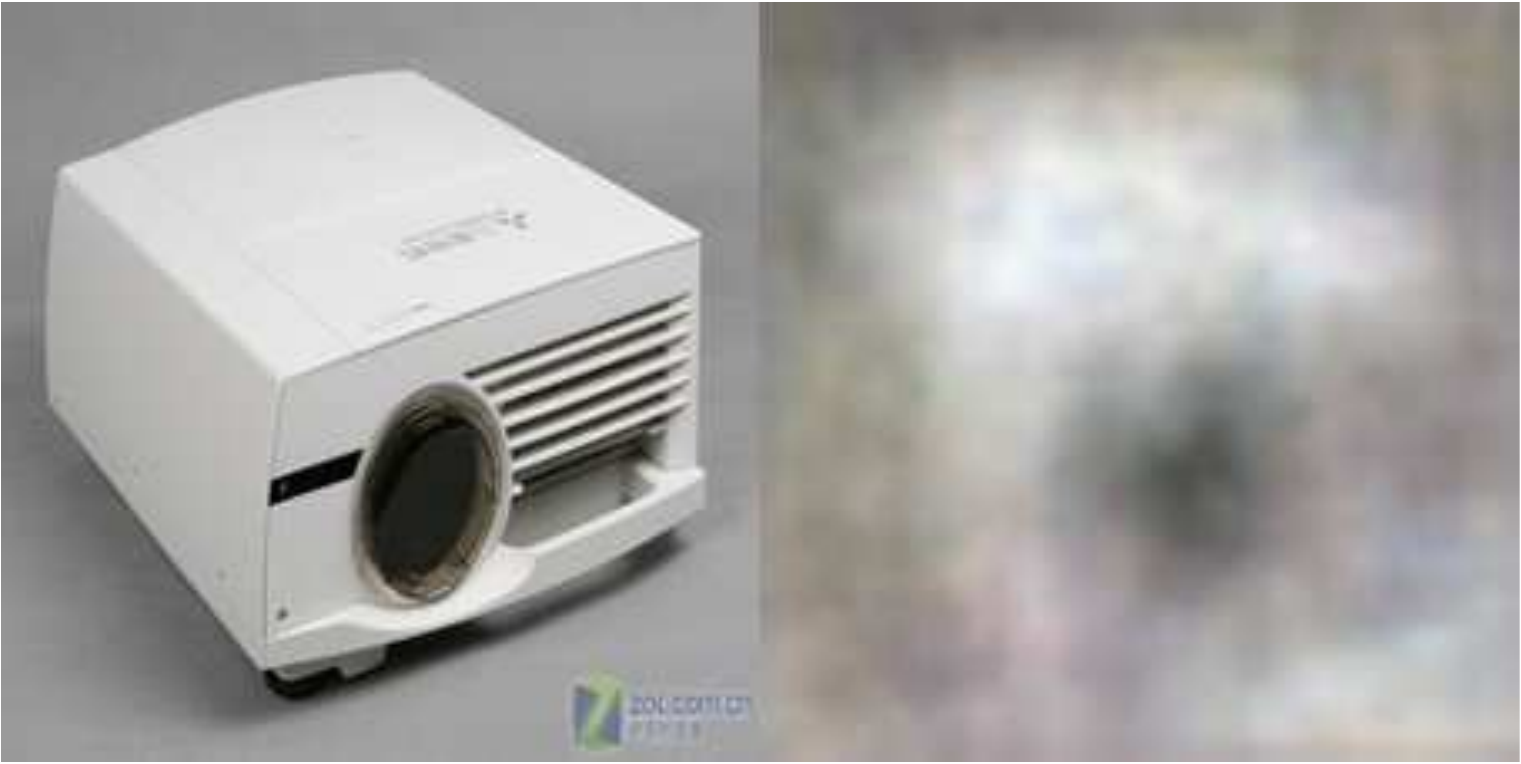}
    &
    \includegraphics[width=\pwid]{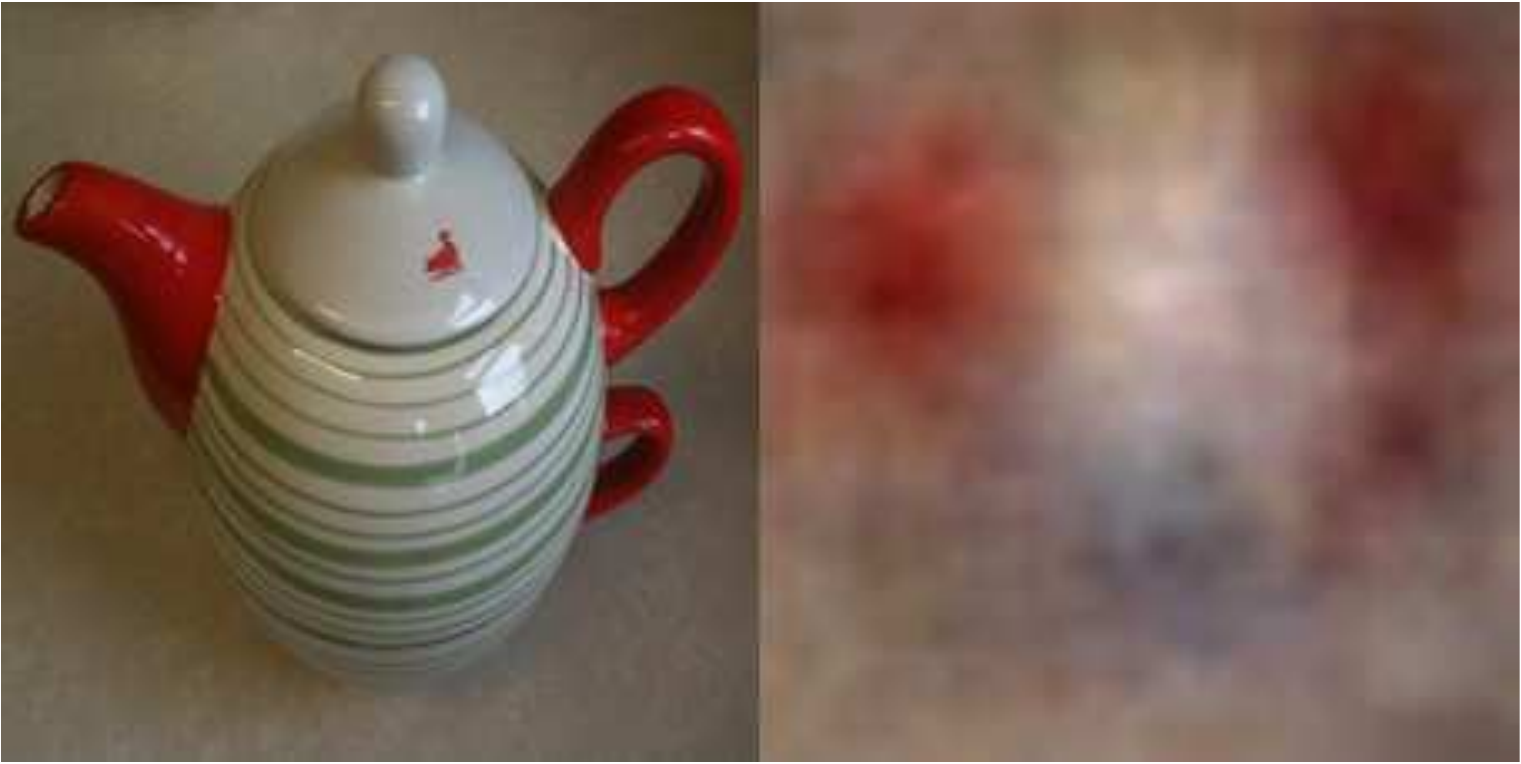}
    &
    \includegraphics[width=\pwid]{Imagenet_RGB/imagenet_13-crop}
    &
    \includegraphics[width=\pwid]{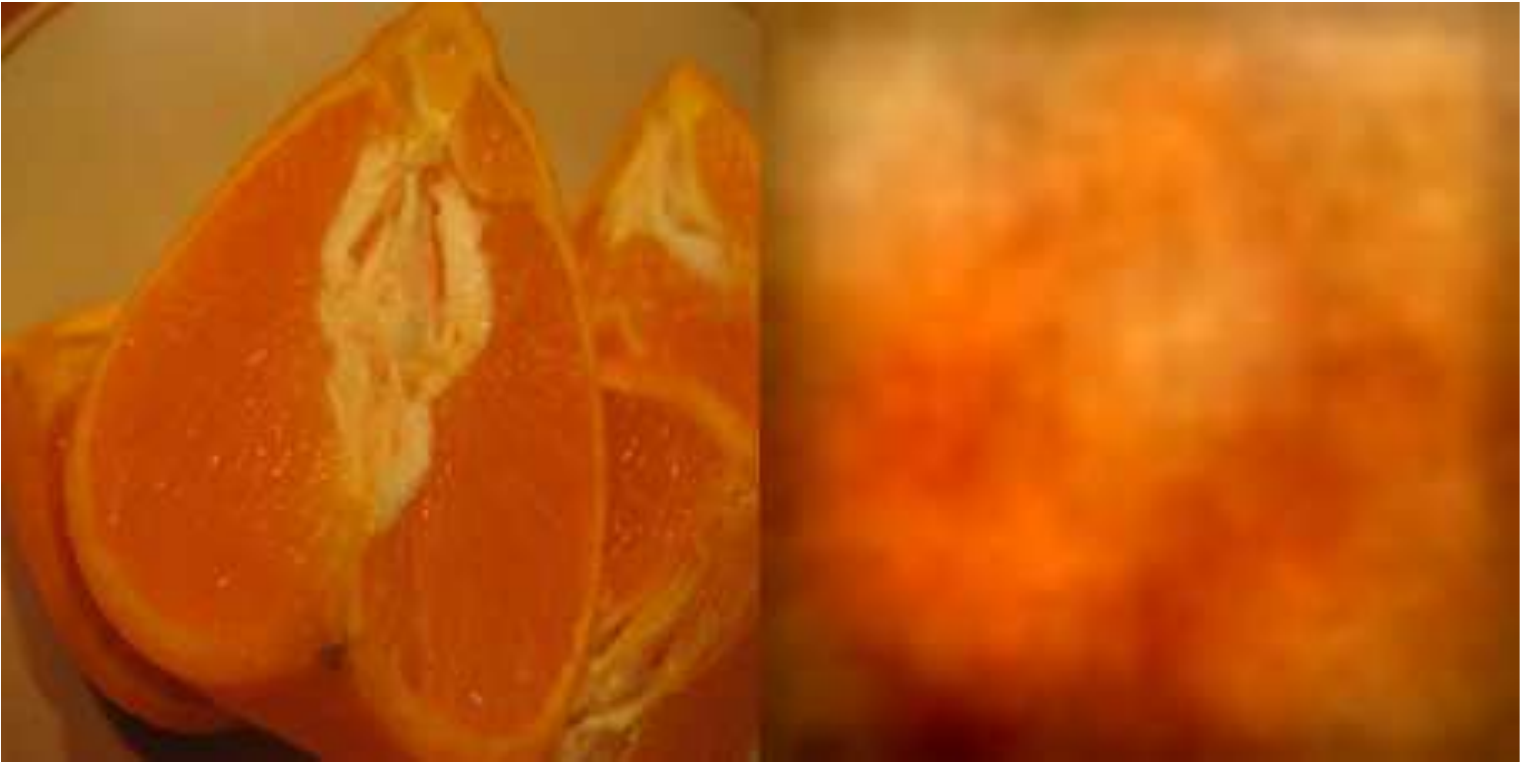}
    &
    \includegraphics[width=\pwid]{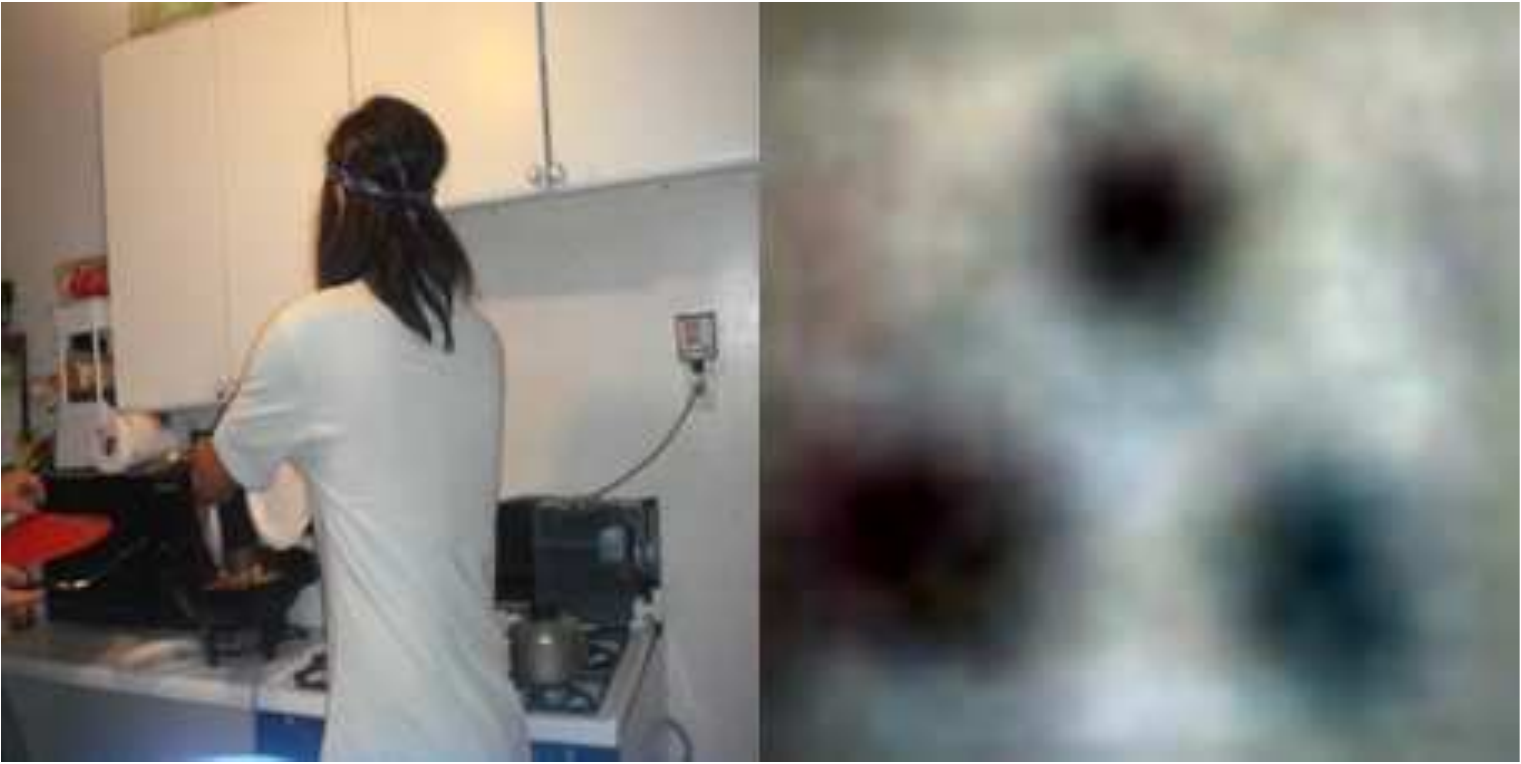}\\
        \includegraphics[width=\pwid]{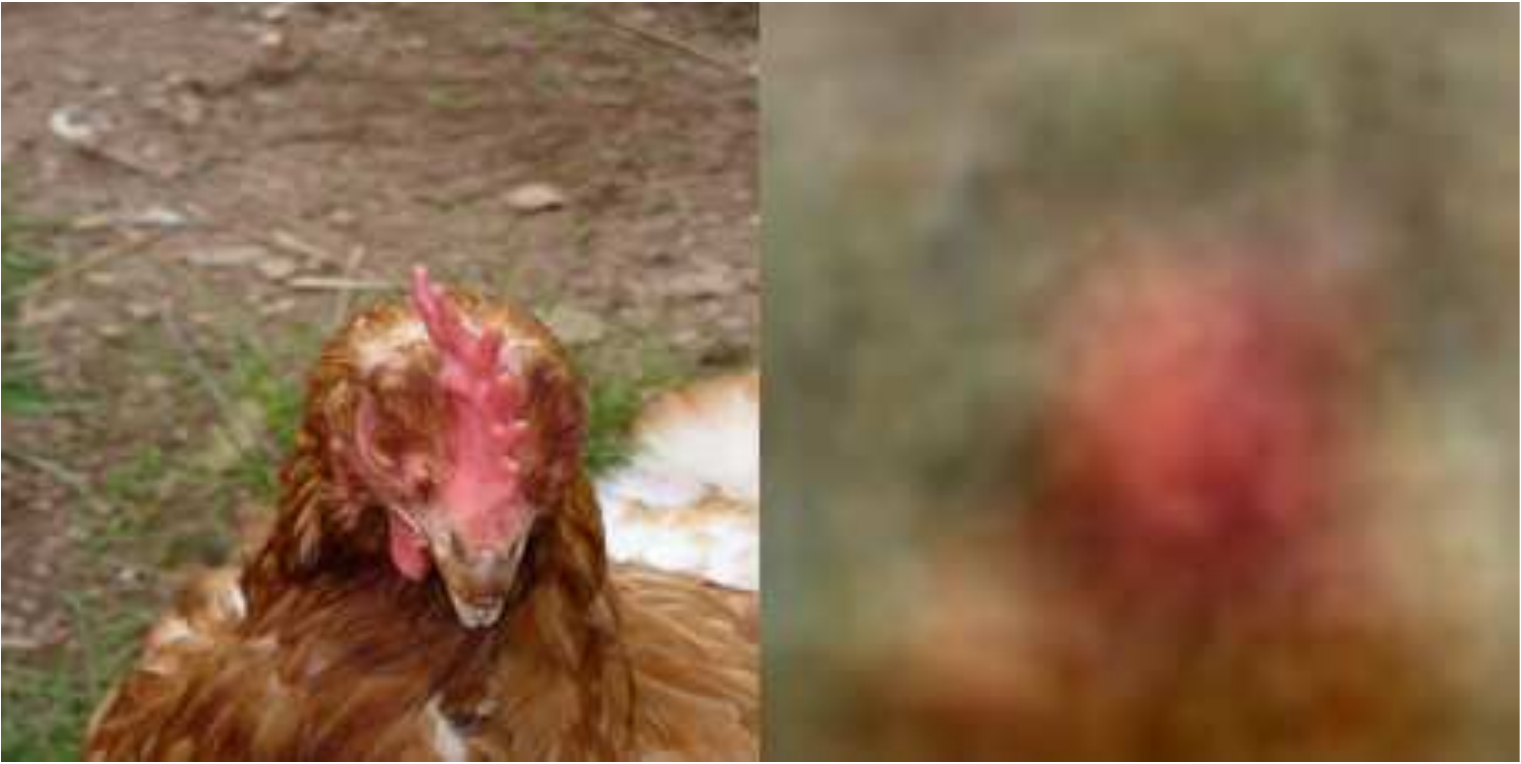}
    &
    \includegraphics[width=\pwid]{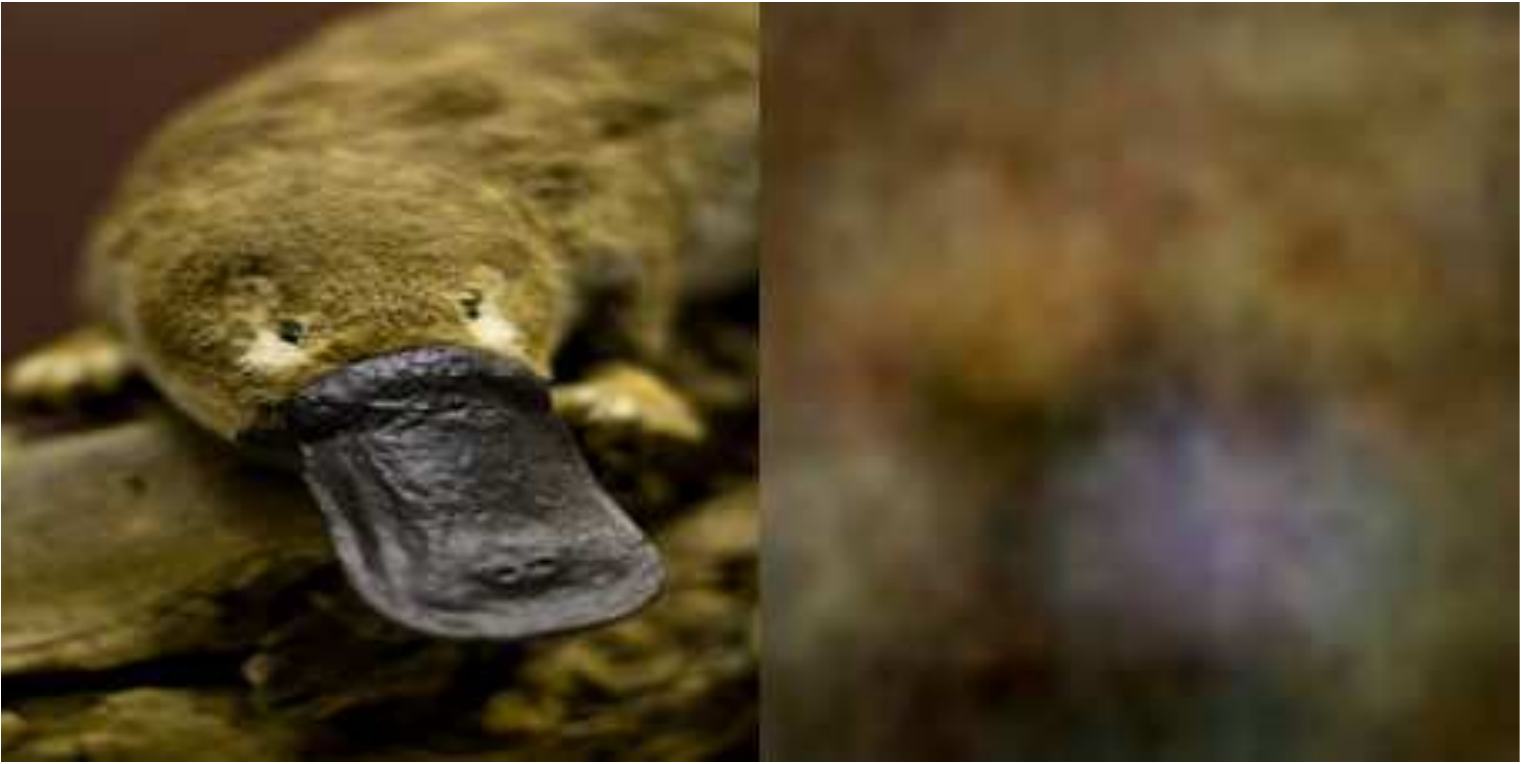}
    &
    \includegraphics[width=\pwid]{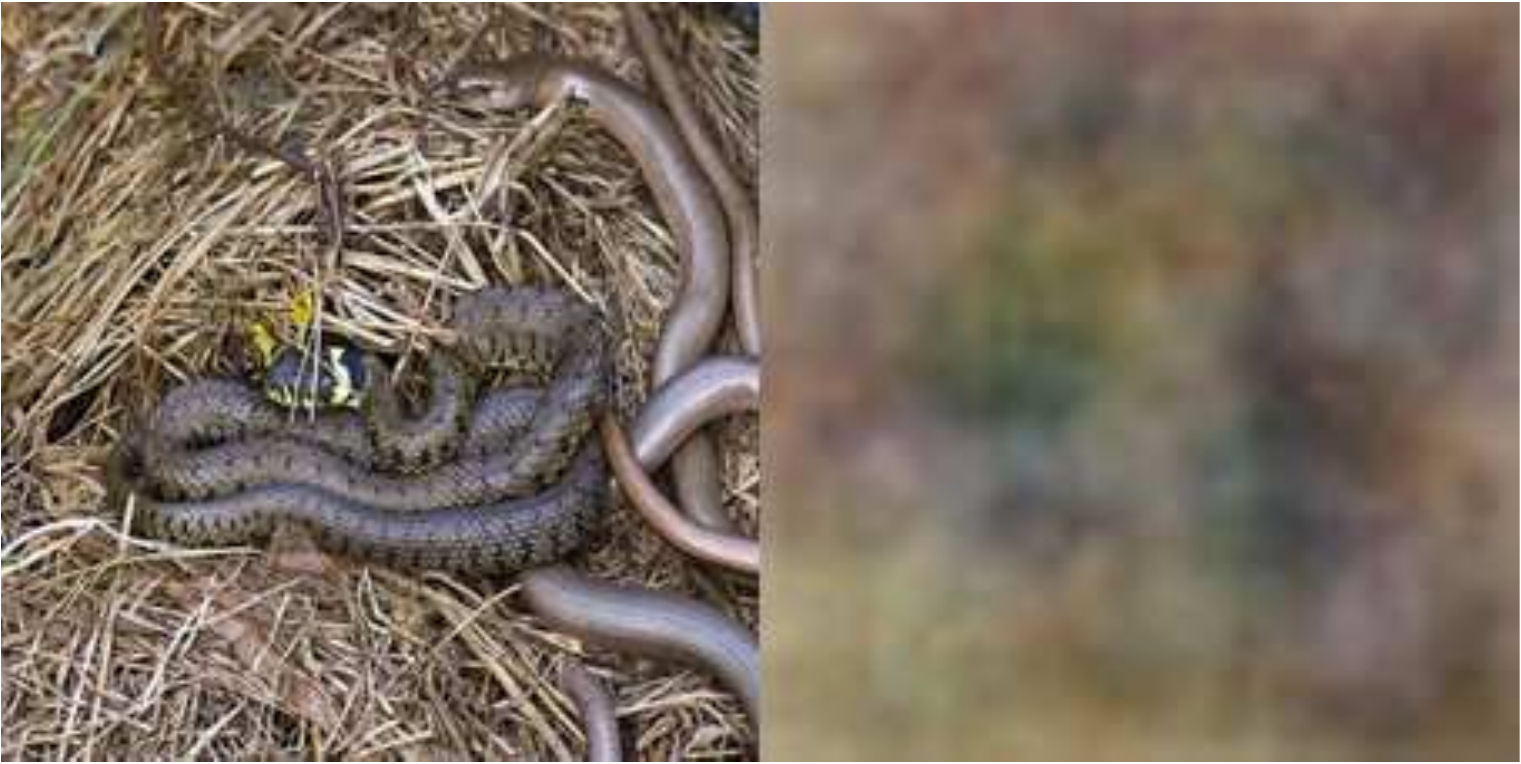}
    &
    \includegraphics[width=\pwid]{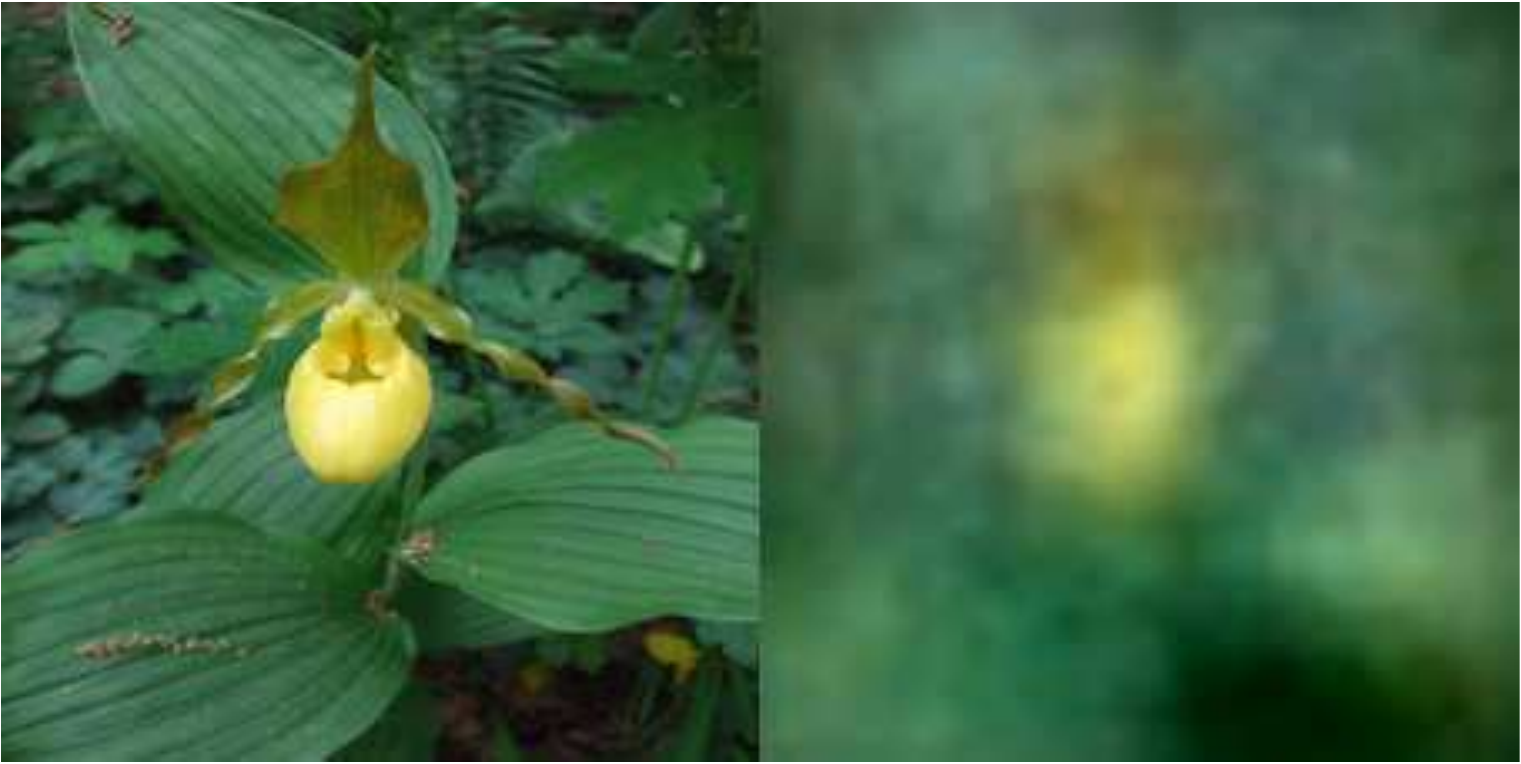}
    &
    \includegraphics[width=\pwid]{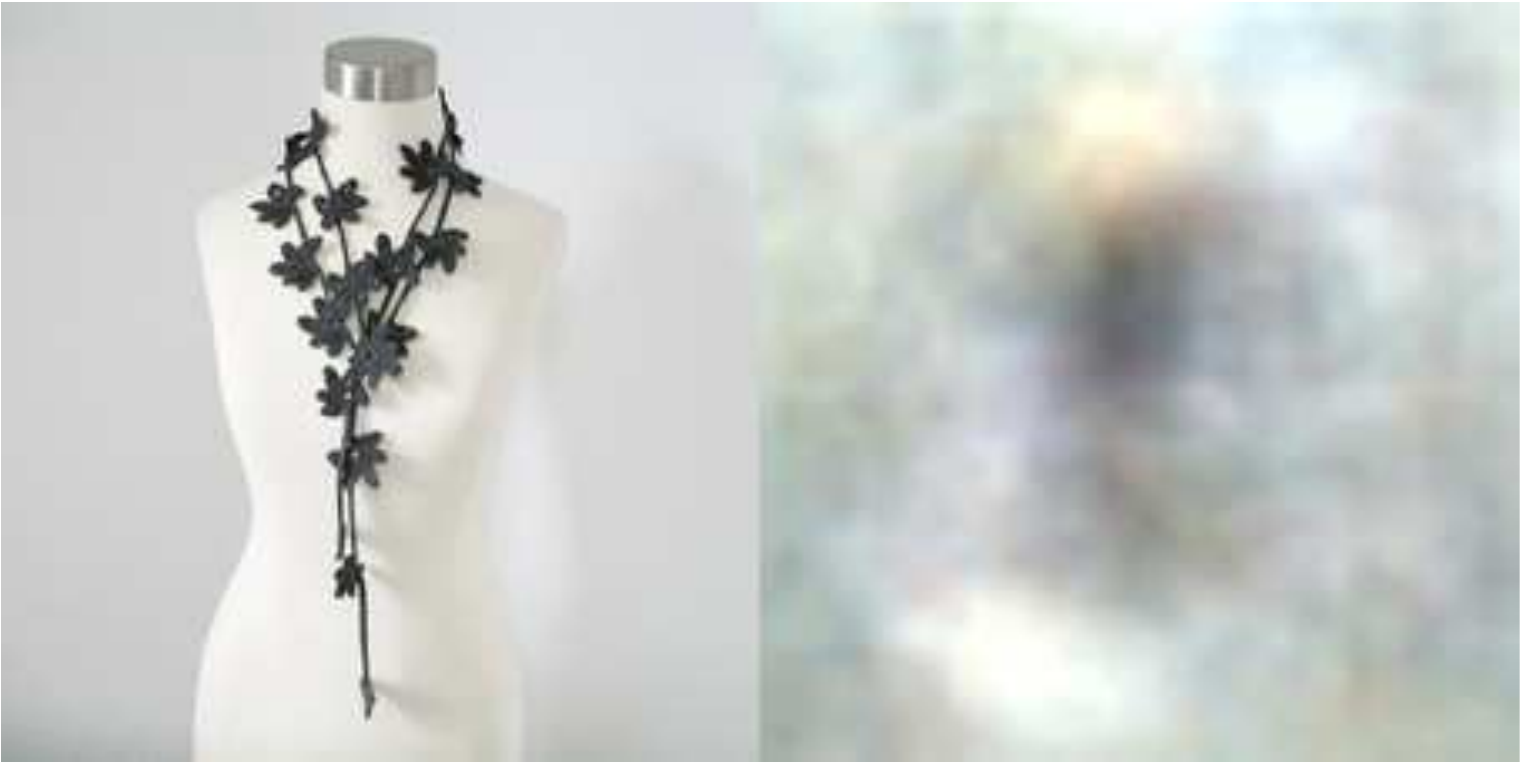}\\
        \includegraphics[width=\pwid]{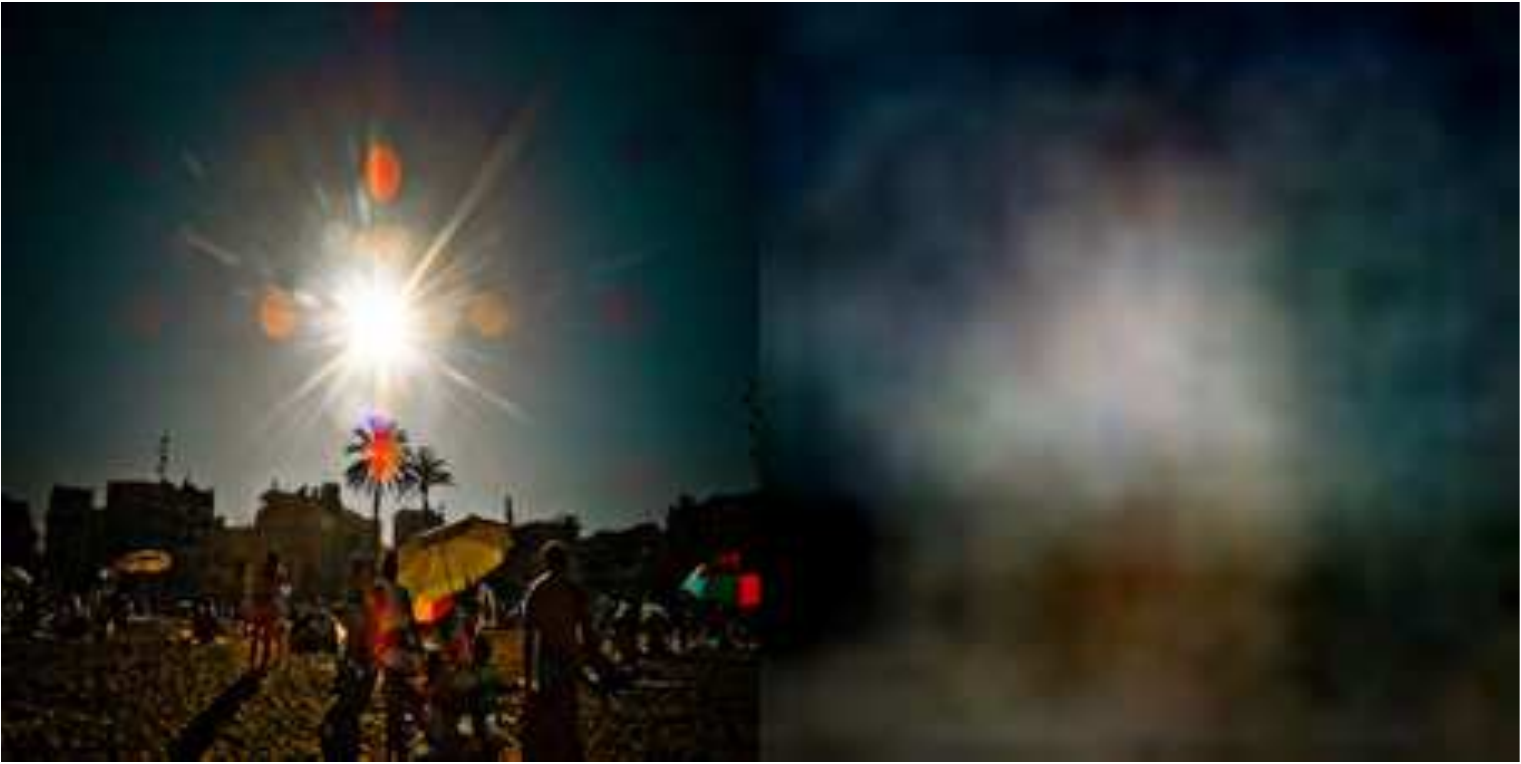}
    &
    \includegraphics[width=\pwid]{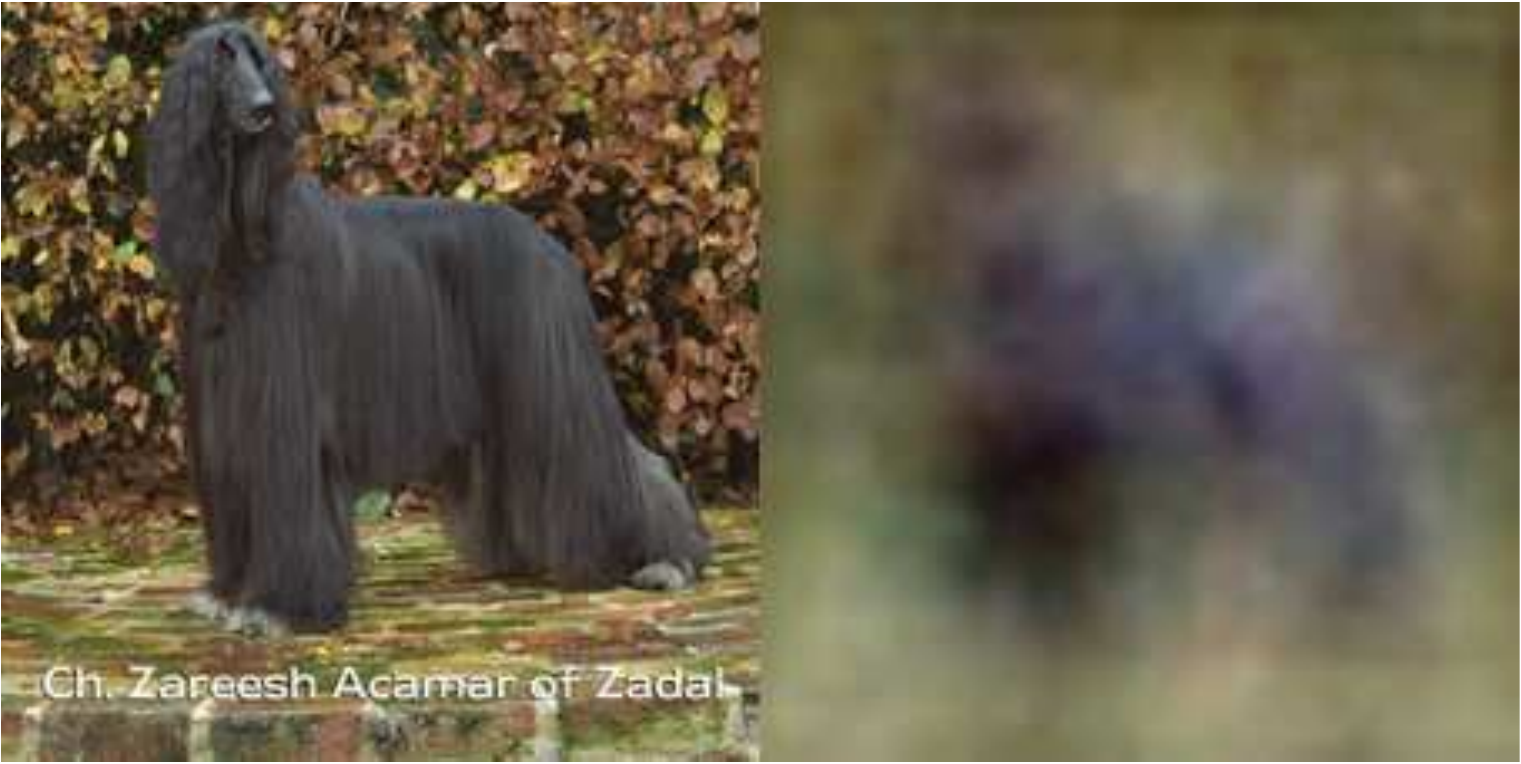}
    &
    \includegraphics[width=\pwid]{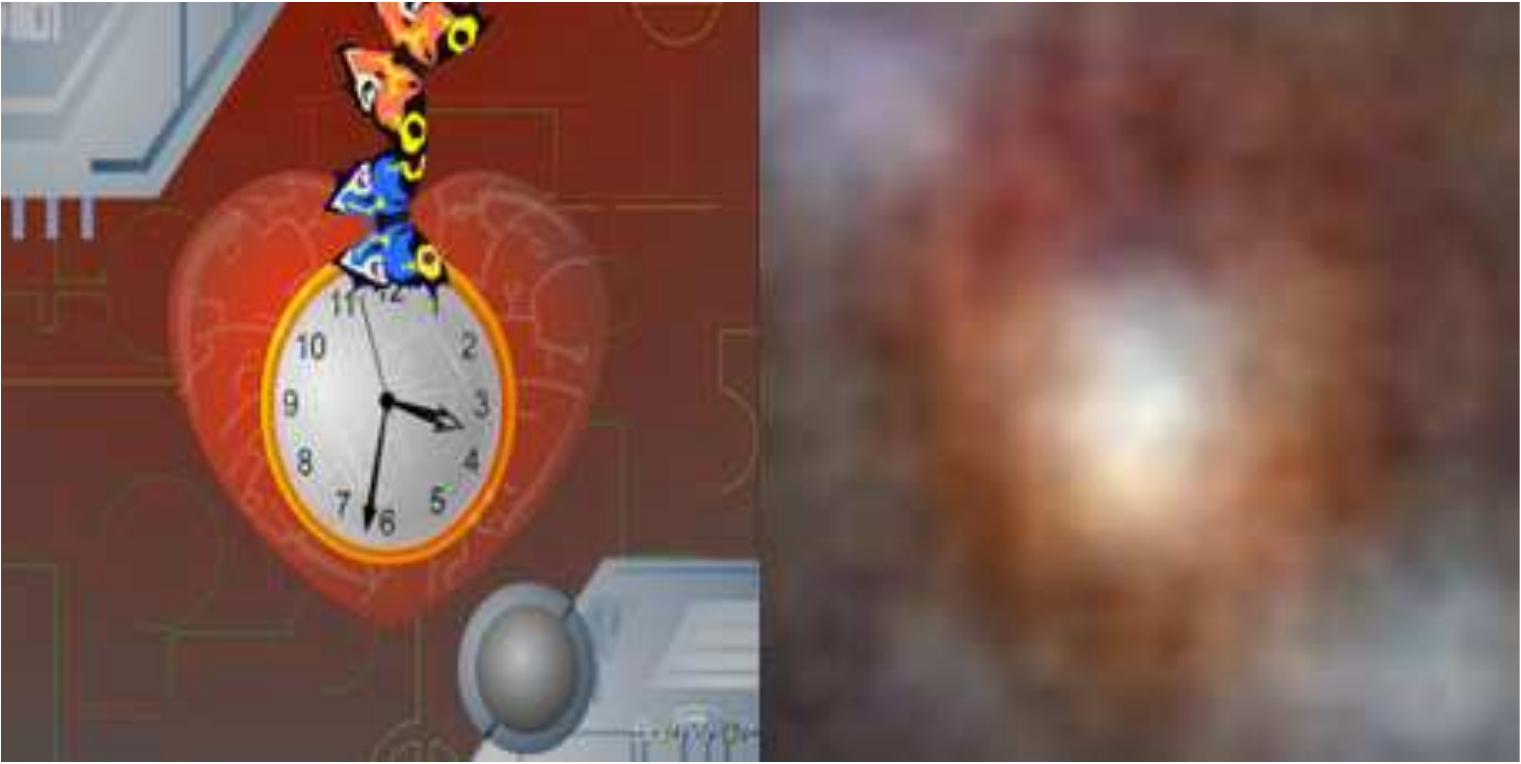}
    &
    \includegraphics[width=\pwid]{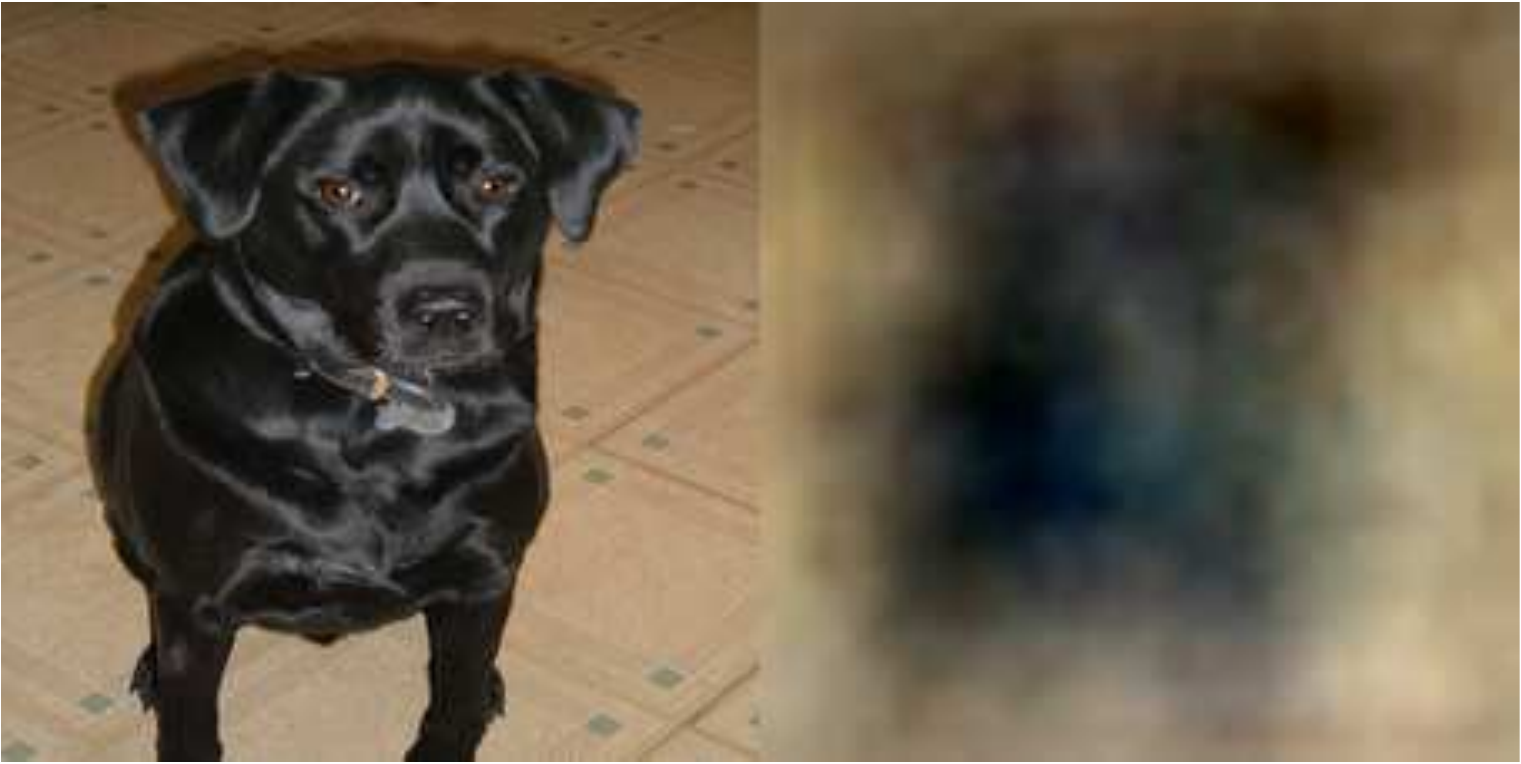}
    &
    \includegraphics[width=\pwid]{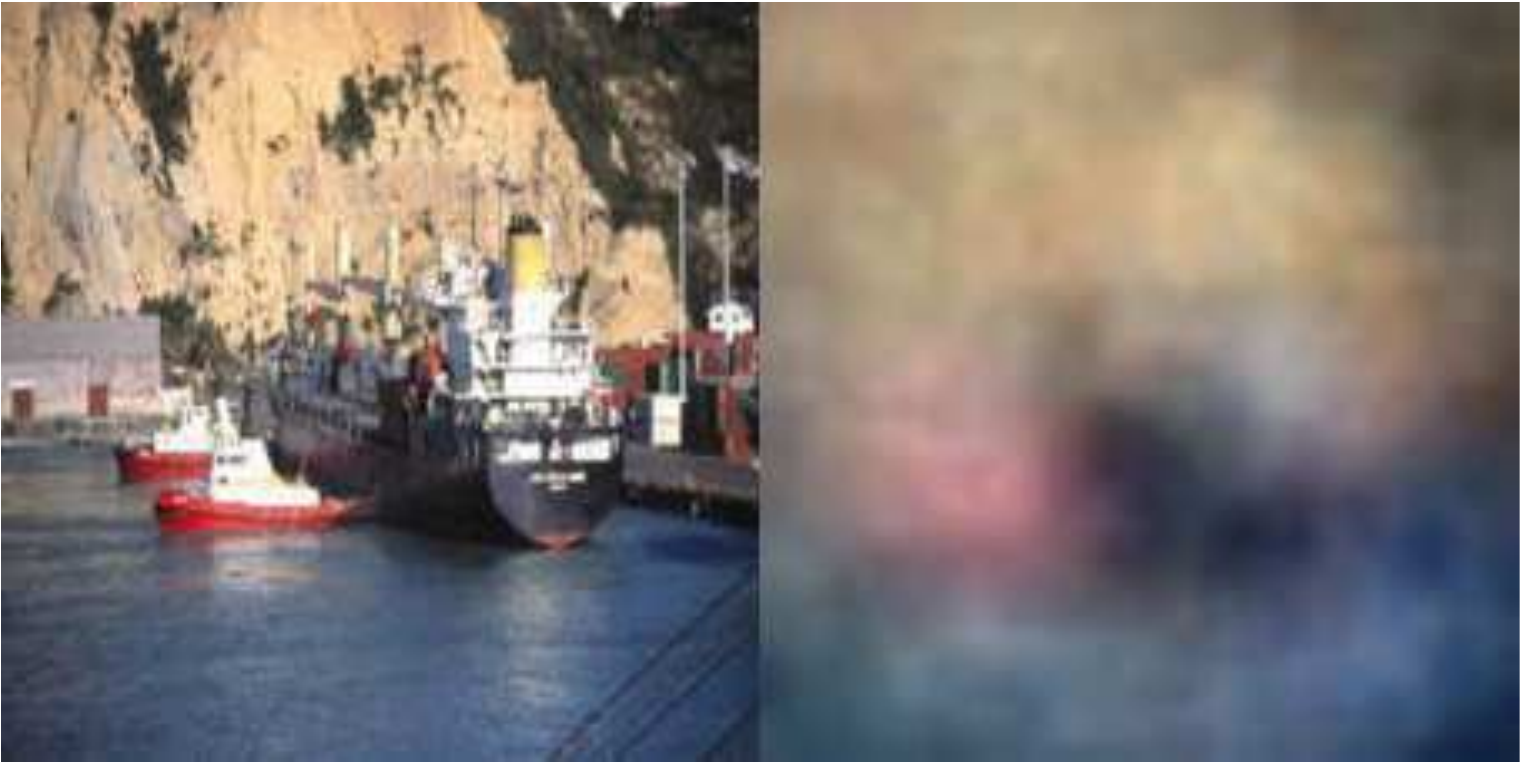}
  \end{tabular}
\caption{\small RGB information linearly predicted from the \CNN{}
  representation. For each pixel we train 3 independent linear
  regressors to predict the pixel's RGB value from the image's global
  \CNN{} representation. We used the first 49K images from ImageNet's
  cross validation set for training and visualized the result for the
  last 1k images. Shown above are the results for 25 random images
  (left) taken from the test set and their reconstructions (right).}
\label{fig:imagenet_reconstruction}
\end{figure*}

Table \ref{tab:VOC11_keypoint_result} reports the accuracy of our
results for keypoint prediction, together with those achieved by other
methods \cite{Long:arxiv:14}. The accuracy is measured using mean PCK
\cite{Yang:pami:13}. A keypoint is considered to be correctly
estimated if the prediction's Euclidean distance from the ground truth
position is $\alpha \in [0,1]$ times the maximum of the bounding box
width and height. Our simple approach outperforms SIFT by a huge
margin on localizing landmarks and is only slightly below the
performance of SIFT+prior \cite{Long:arxiv:14}.  Figure
\ref{fig:keypoint_PCK_alpha} shows the plot of PCK vs $\alpha$ for 20
classes.

\subsubsection{RGB reconstruction}
\label{sec:rgb_reconstruction}
The results from the previous tasks show that our \CNN{}
representation does encode some levels of spatial information. The natural question is then \emph{what does it actually remove?} To investigate this we try to evaluate if it is possible to invert the \CNN{} mapping.
First, we try to estimate the RGB values
of the original input image from our \CNN{} representation. For this,
again, we simply learn $3 \times n_p$ linear regressors where $n_p$ is
the number of pixels in the image. In other words we learn an
independent regressor for each pixel and each colour channel.

We use ImageNet as our testbed.  We used the first 49k images from
ImageNet's validation set for training and the last 1k images for
testing.  We resized each image to 46$\times$46$\times$3 and trained
6348 independent linear regressors.  Some examples of the resulting
RGB reconstruction are illustrated in figure
\ref{fig:imagenet_reconstruction}.  The mean absolute error of image
reconstruction is 0.12.  It is rather surprising that RGB values of an
image can be extracted with this degree of accuracy from the \CNN{}
representation.



\subsubsection{Semantic Segmentation}
\label{sec:sem_seg}

\begin{table*}[]
  \scriptsize
  \centering
  \setlength{\tabcolsep}{0.6pt}
\begin{tabular}{@{}l *{22}{c@{\hspace*{5pt}}}}
\toprule
 & background & airplane & bike & bird & boat & bottle & bus & car & cat & chair & cow & table & dog & horse & mbike & person & plant & sheep & sofa & train & tv & \textbf{mean} \\
\midrule
\CNN{} &  79.12 & 16.01  &  0.02 &   12.93 &   9.26 &   13.69 &   37.29  &  33.75 &   40.01 &   0.01 &   8.62
&    12.24  &  30.89 &   9.43 &   24.94  &  44.03 &   6.22  &  18.77 & 1.64  &  25.33  & 11.05 & 20.73 \\
\bottomrule
\end{tabular}
\vspace{0.1cm}
\caption{\small Evaluation of Semantic Segmentation on the
  validation set of VOC12 measured in mean Average Precision (mAP).}
\vspace{0.3cm}
\label{tab:VOC12_semantic_seg}
\end{table*}

\def\pwid{0.318\linewidth}
\begin{figure*}[t!]
  \centering
  \begin{tabular}{@{}ccc@{}}
    \includegraphics[width=\pwid]{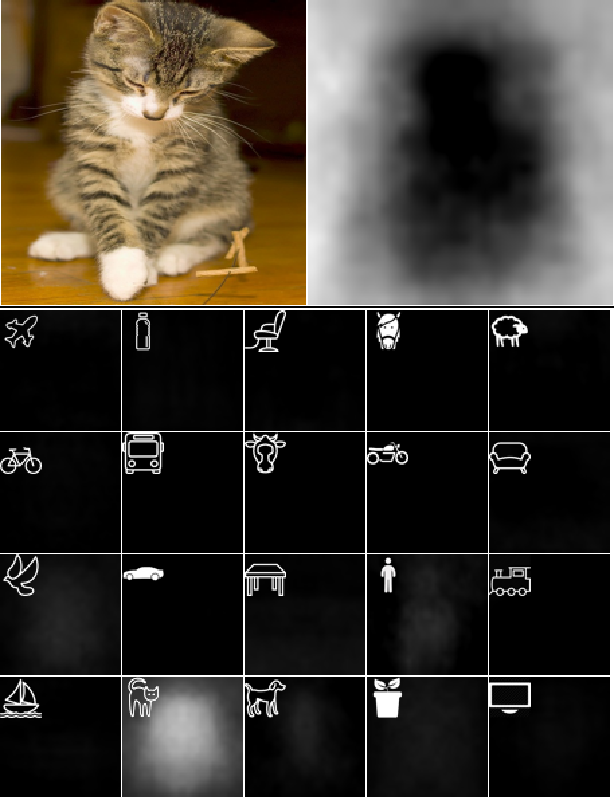} &
    \includegraphics[width=\pwid]{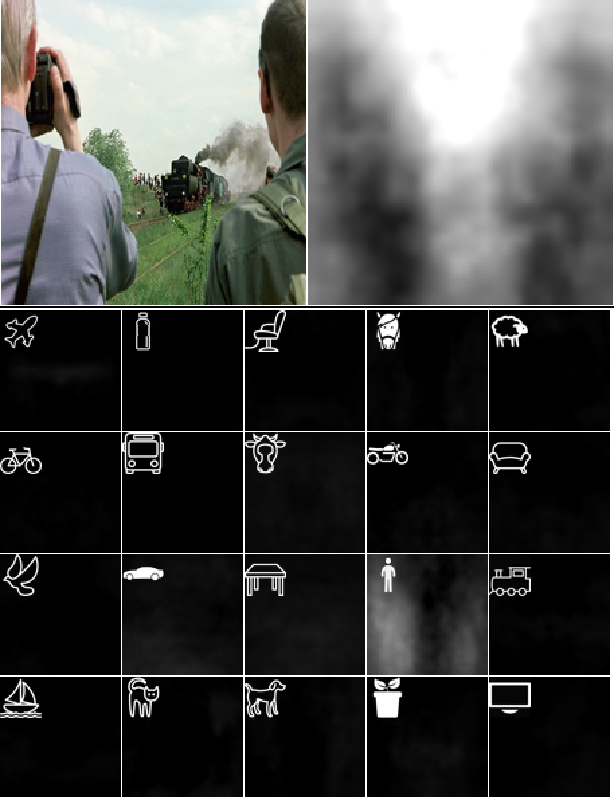} &
    \includegraphics[width=\pwid]{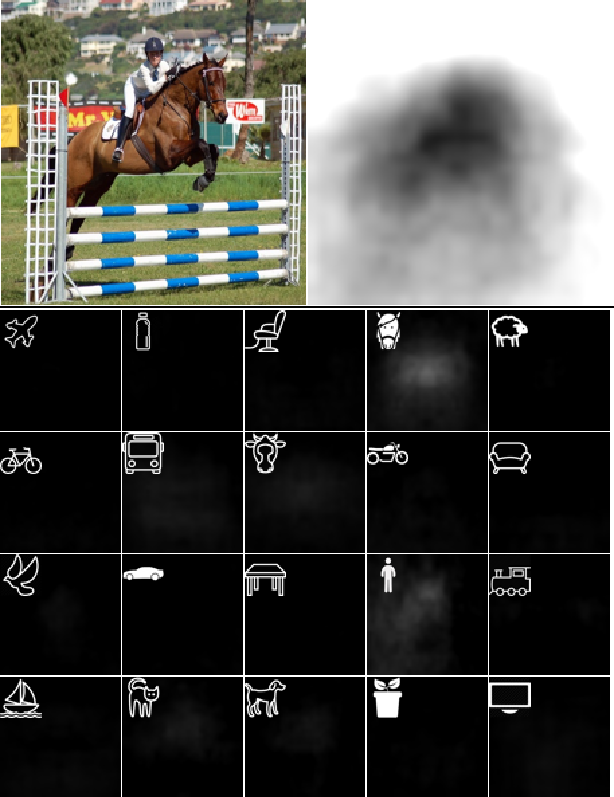}\\
    \includegraphics[width=\pwid]{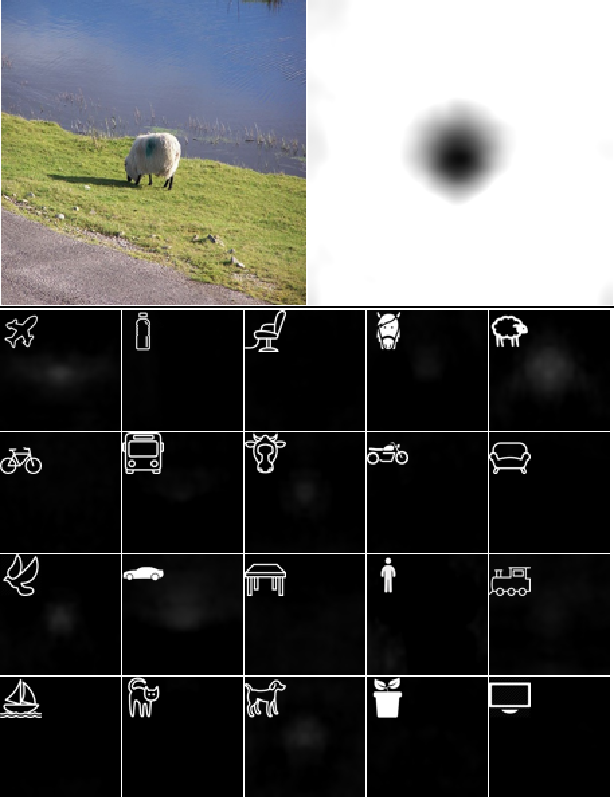} &
    \includegraphics[width=\pwid]{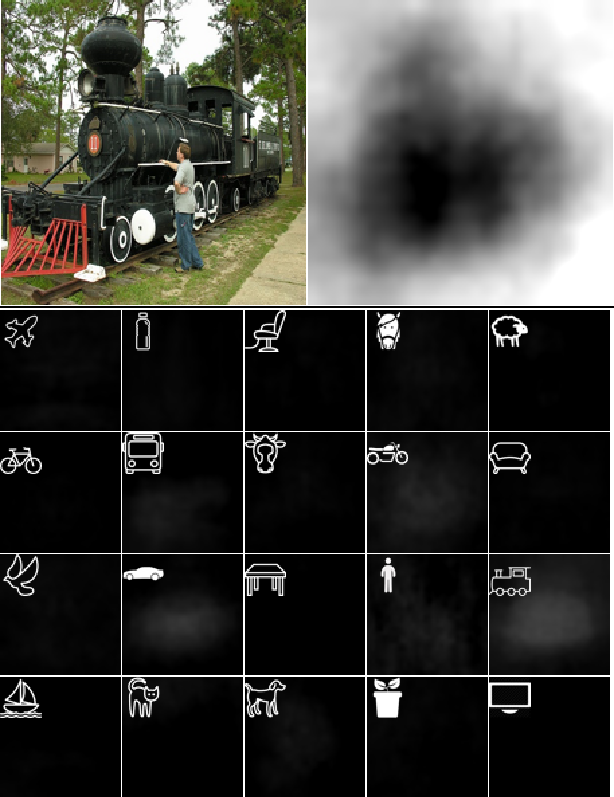} &
    \includegraphics[width=\pwid]{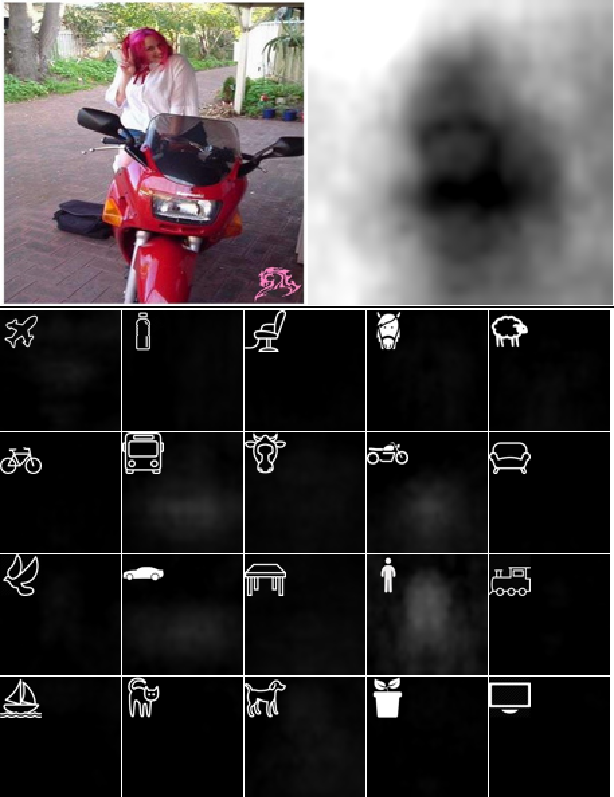}\\
  \end{tabular}
  \caption{\small Semantic segmentation results for images from PASCAL
    VOC. For each block of pictures, the top left hand picture is the
    original image and the one directly to its right is the
    \textit{probability} map for the background class. The brighter
    the pixel the higher the probability. The bottom set of smaller
    images in the block display, in the same manner, the probabilities
    for the 20 classes of PASCAL VOC 2011. The probability masks are
    computed independently of one another though the scaling of the
    intensities in the displayed masks is consistent across all the
    masks. The learning is based on linear regression, the details of
    which are found in the main text.}
  \label{fig:semantic_seg}
\end{figure*}

We applied the same framework which we employed for RGB reconstruction
further to recover semantic labels of each pixel instead of its RGB
values. The procedure is as follows: we resized each semantic
segmentation map of VOC 2012 segmentation task down to a
30$\times$30$\times$21 image. We train a separate linear regressor to
predict whether the pixel at position $(x,y)$ belongs to class $c$ or
not encoded as 1 and 0. We have $x\in\{1, 2, \ldots, 30\}$ and
similarly for $y$ and $c \in\{1, 2, \ldots, 21\}$.  Therefore a total
of 18900 linear regressors are trained with ridge regression.  Solving
a classification problem via regression is not ideal. But the qualitative
results shown in figure \ref{fig:semantic_seg} are visually
pleasing. They show the semantic segmentations produced by our
approach for some images from the VOC12 validation set.

After we apply the linear regressor for each class to each pixel, we
get 21 responses. We then turn these responses into a single
prediction using another linear model.  We multiply the response
vector by a matrix $M \in \mathbb{R}^{21\times21}$ and then choose the
class which corresponds to the highest response in the output
vector. Ideally $M$ should model the relations between different class
responses at a single pixel.  We learn $M$, once again with ridge
regression, and during training it tries to return a binary vector of
length 21 with only one non-zero entry.

As our segmentation masks only have size 30$\times$30 we resize the
them back to their original size. We used the VOC12 training-set as
the training data and augmented this set tenfold to get a better
estimate and reported the result on the cross-validation
set. Quantitative results for our segmentations are given in table
\ref{tab:VOC12_semantic_seg}. Although this result itself is not as
good as s.o.a. on semantic segmentation task (mean average precision
of 20.7 compared to 47.5 of s.o.a. method), it is intriguing to see
that the global \CNN{} representation contains this level of
information. It is easy to envisage that such a segmentation could be
incorporated into an object classifier or detector.

%% file: Feature_Extrapolation.tex
\def\pwid{.12\linewidth}
\begin{figure}[h!]
  \centering
  \begin{tabular}{@{}*{7}{c@{\hspace*{5pt}}}}
            \multicolumn{3}{l}{\small male} &
            \multicolumn{1}{c}{$\longrightarrow$} &
            \multicolumn{3}{r}{\small female}\\   
    \includegraphics[width=\pwid]{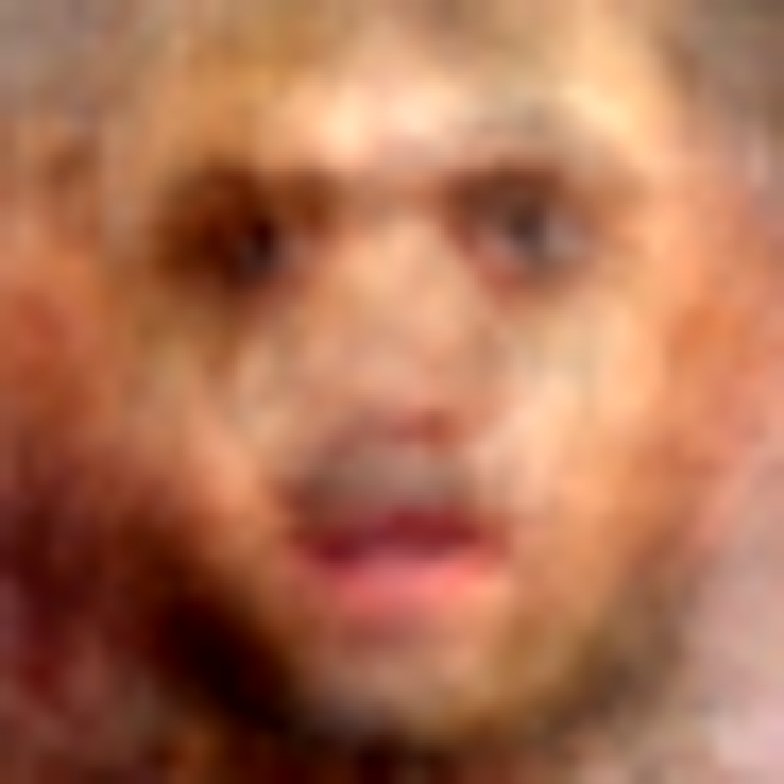} &
    \includegraphics[width=\pwid]{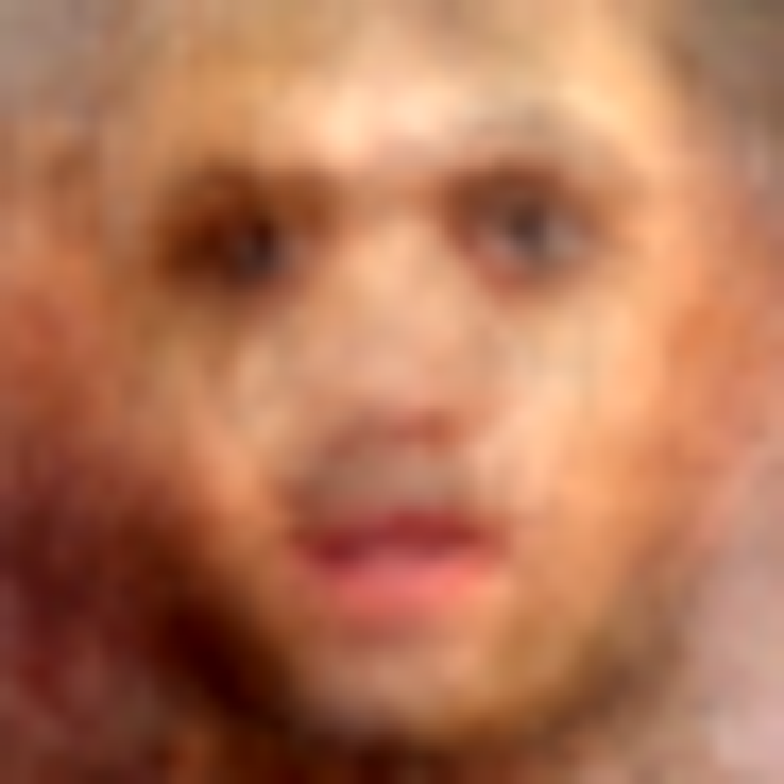} &
    \includegraphics[width=\pwid]{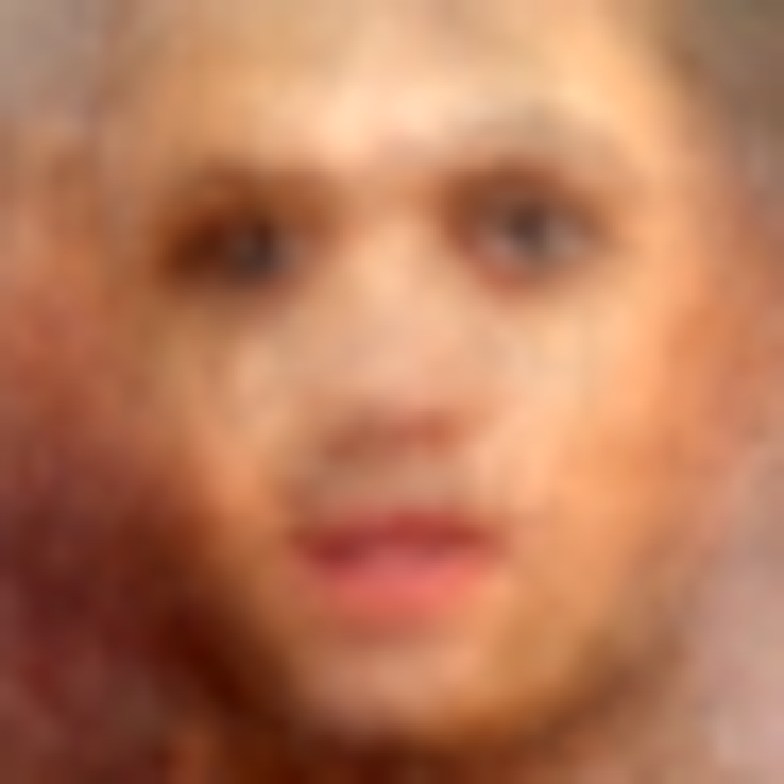} &
    \includegraphics[width=\pwid]{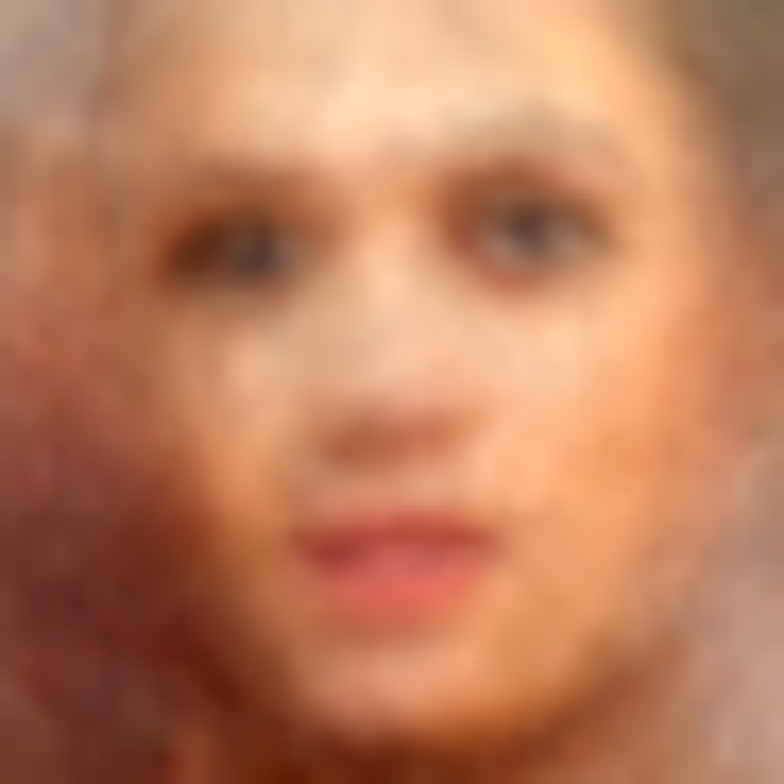} &
    \includegraphics[width=\pwid]{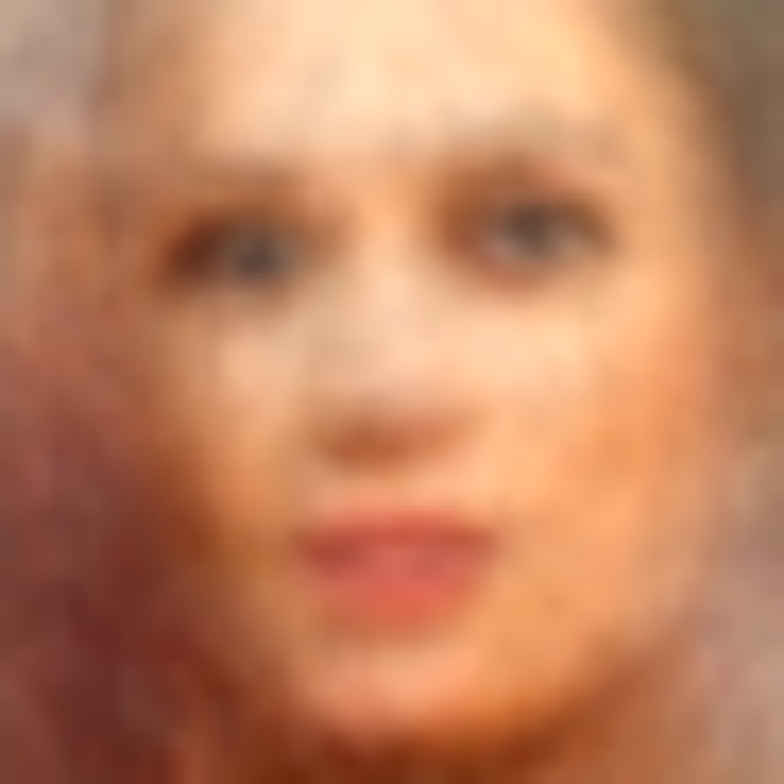} &
    \includegraphics[width=\pwid]{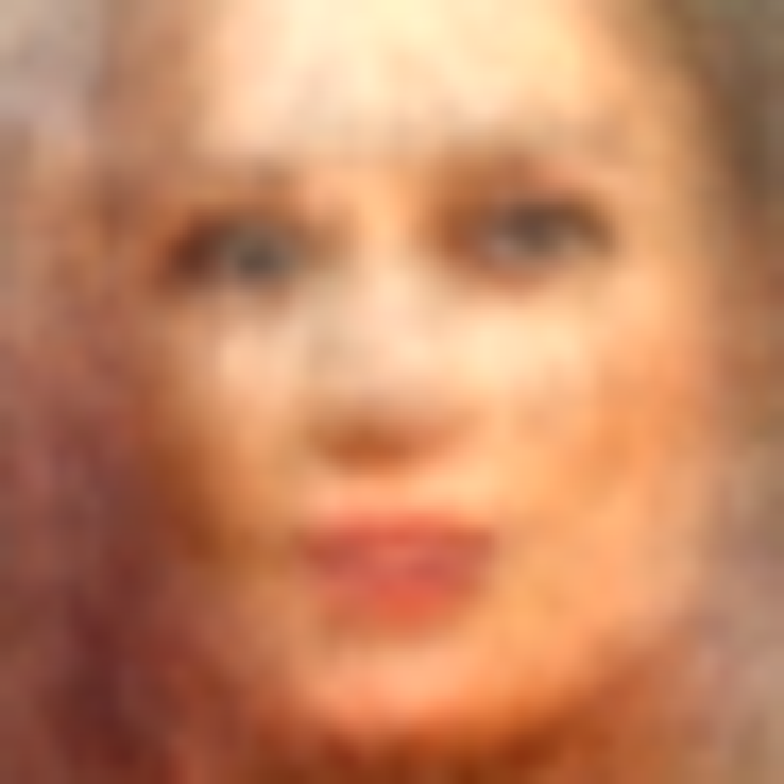} &
    \includegraphics[width=\pwid]{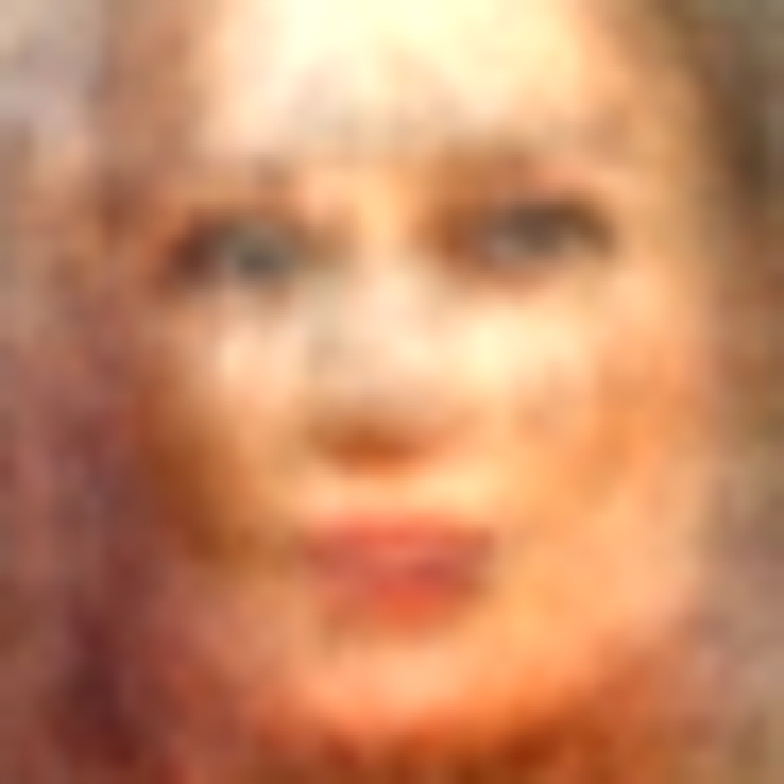}\\[5pt]
                \multicolumn{3}{l}{\small no glasses} &
            \multicolumn{1}{c}{$\longrightarrow$} &
            \multicolumn{3}{r}{\small glasses}\\   
    \includegraphics[width=\pwid]{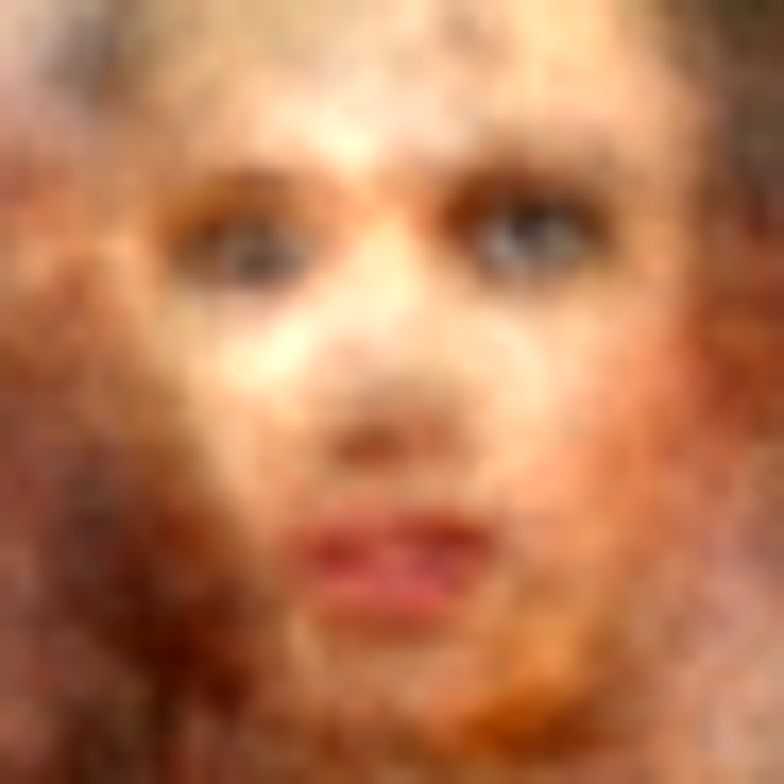} &
    \includegraphics[width=\pwid]{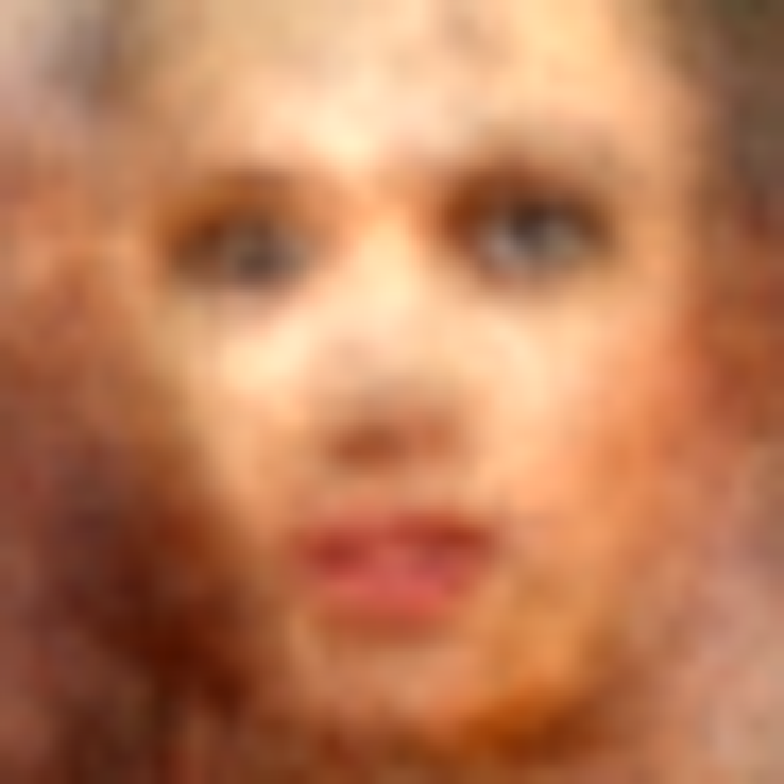} &
    \includegraphics[width=\pwid]{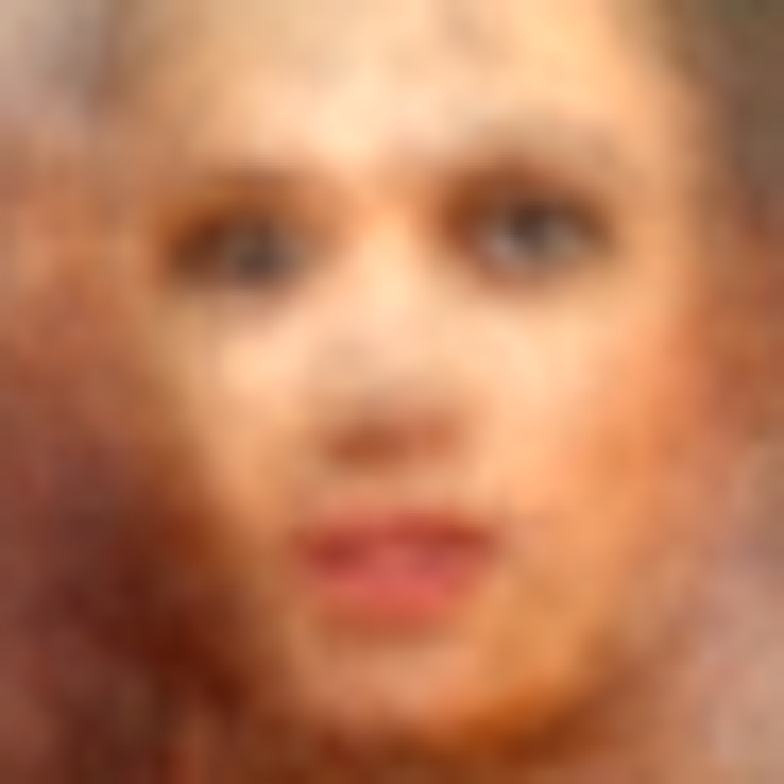} &
    \includegraphics[width=\pwid]{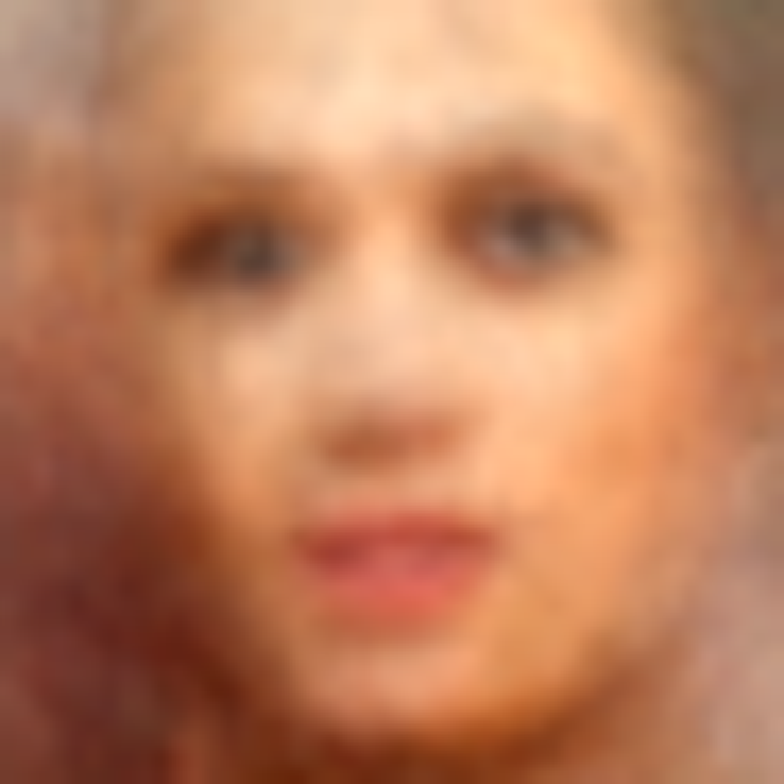} &
    \includegraphics[width=\pwid]{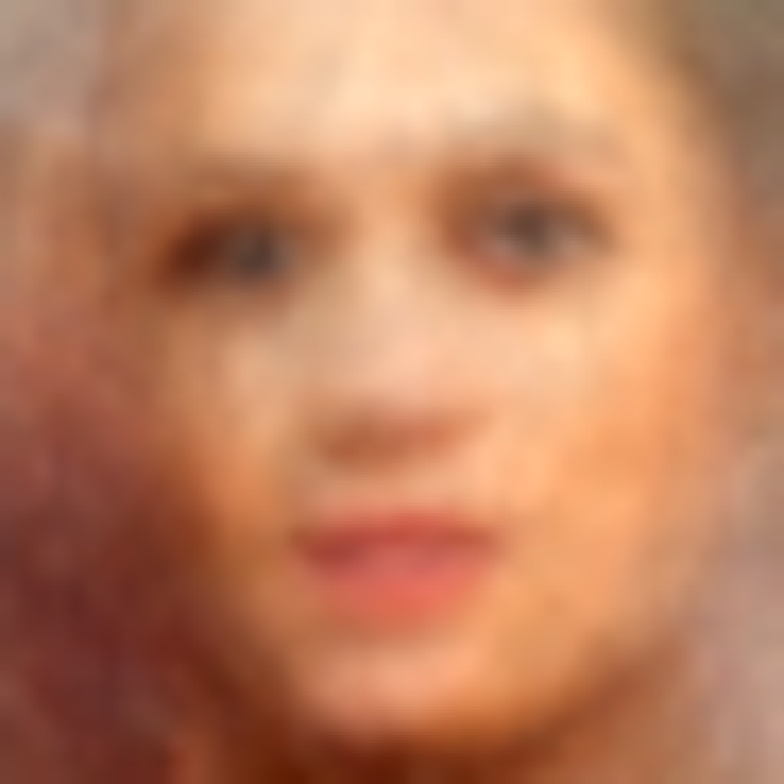} &
    \includegraphics[width=\pwid]{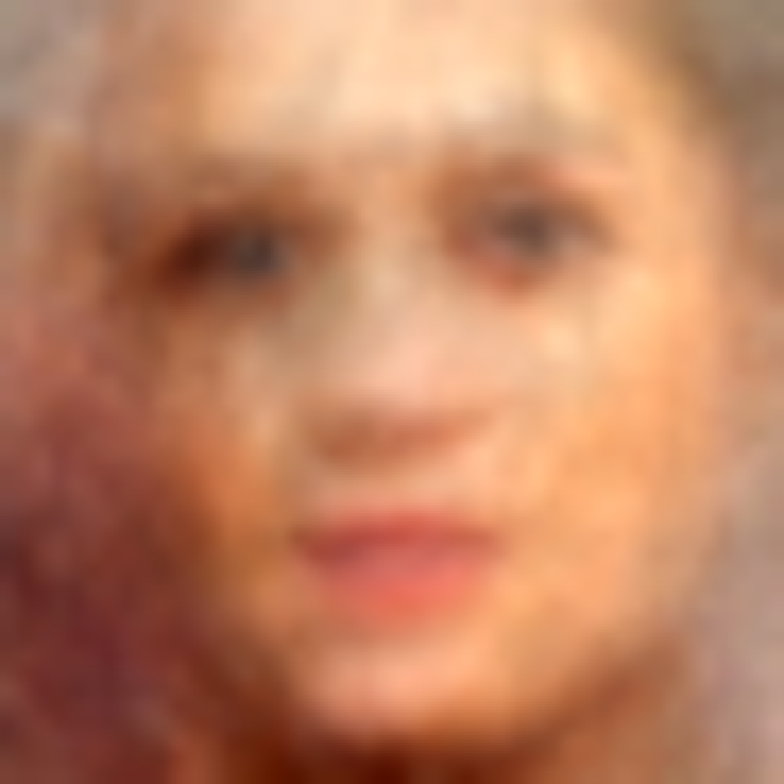} &
    \includegraphics[width=\pwid]{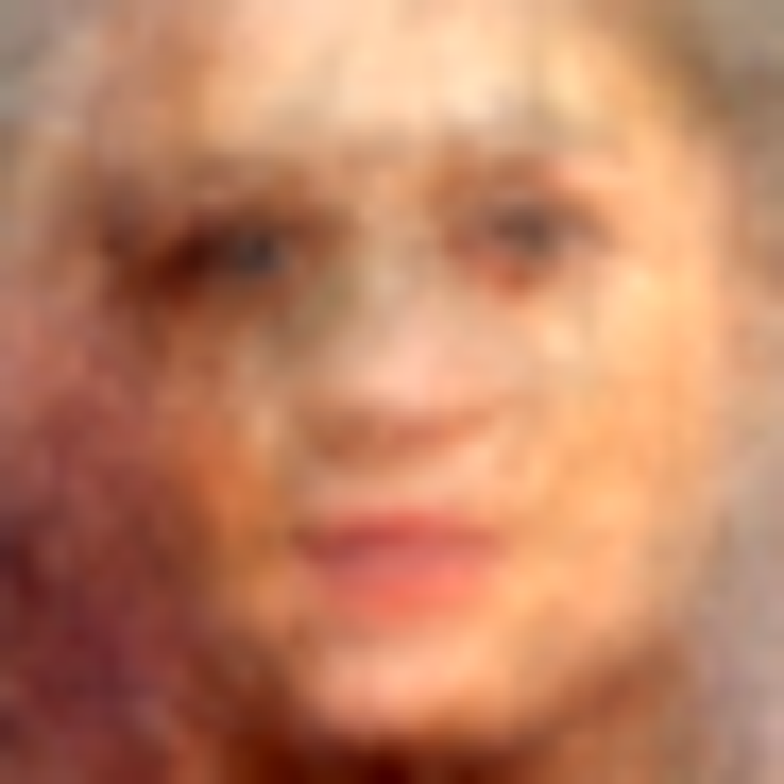}\\[5pt]
                    \multicolumn{3}{l}{\small right in-plane rotation} &
            \multicolumn{1}{c}{$\longrightarrow$} &
            \multicolumn{3}{r}{\small left in-plane rotation}\\   
    \includegraphics[width=\pwid]{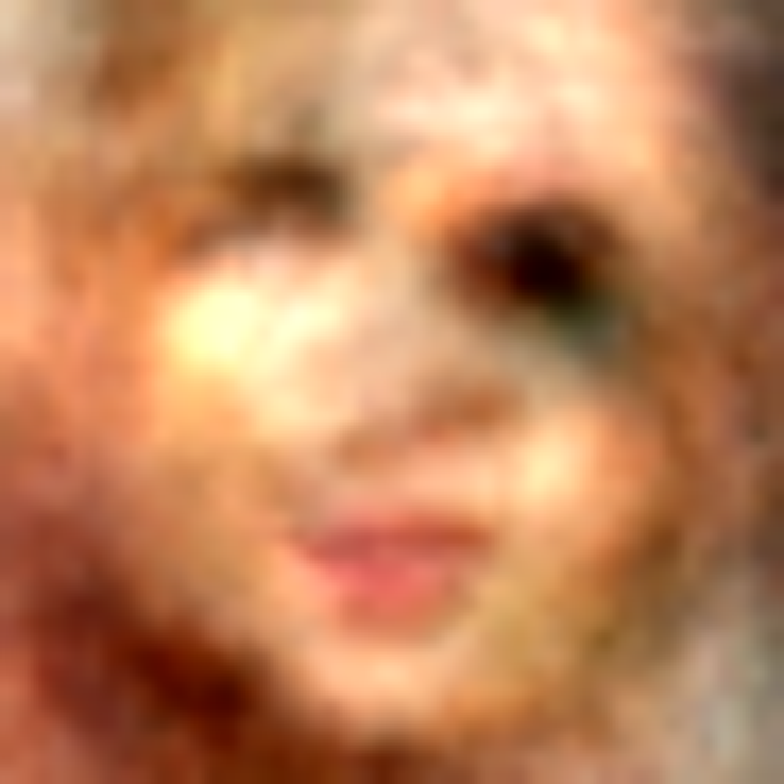} &
    \includegraphics[width=\pwid]{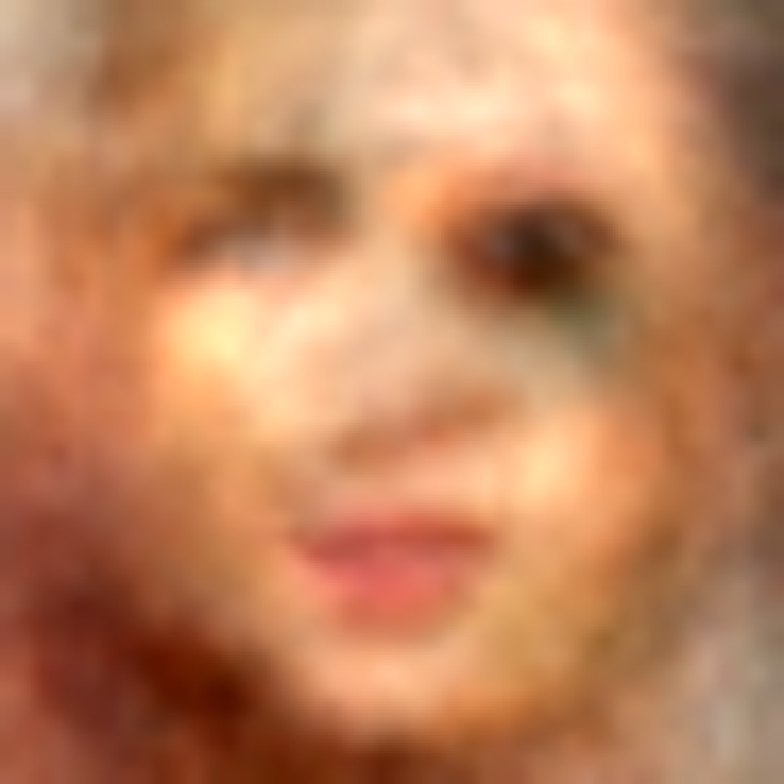} &
    \includegraphics[width=\pwid]{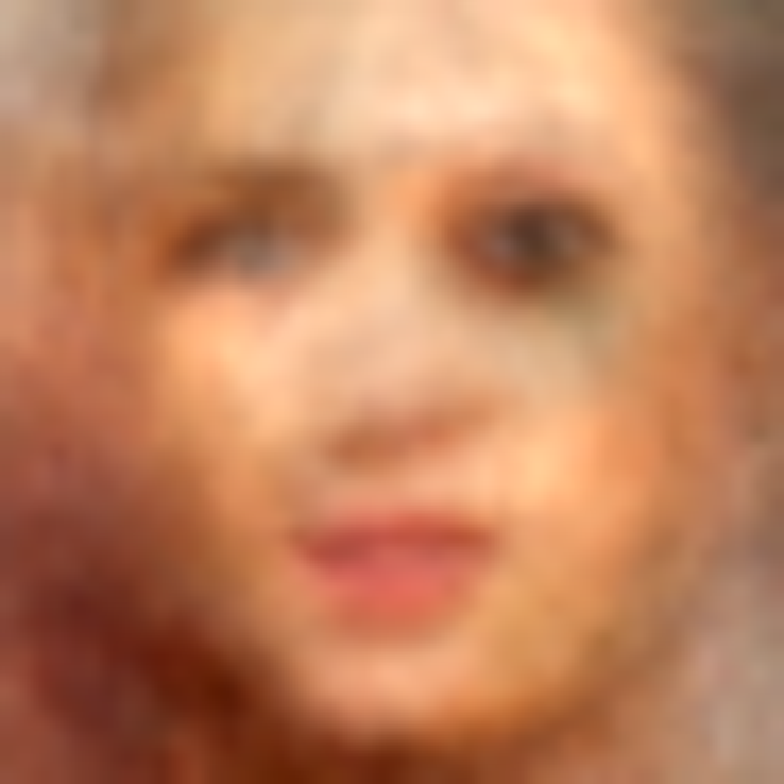} &
    \includegraphics[width=\pwid]{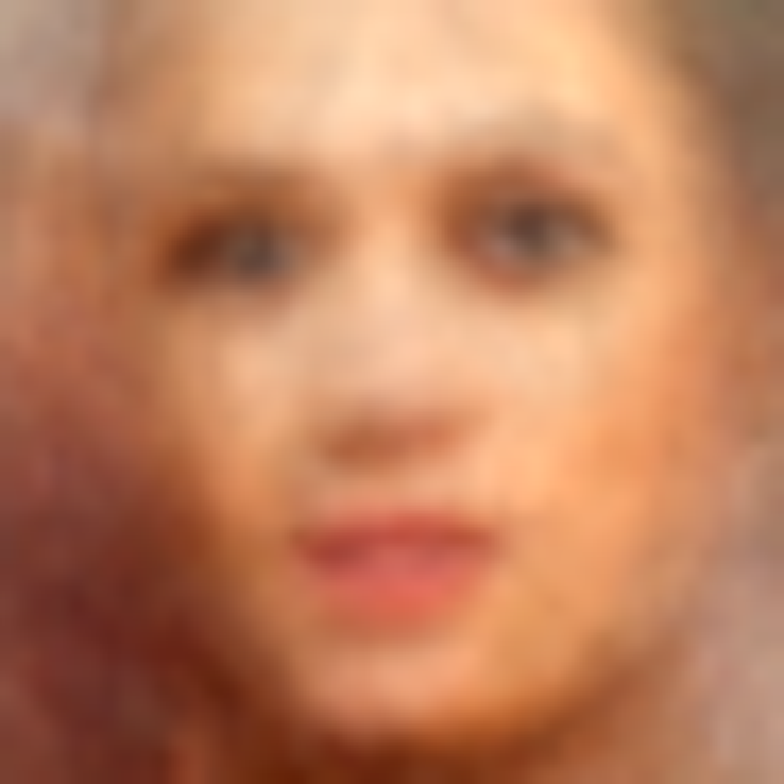} &
    \includegraphics[width=\pwid]{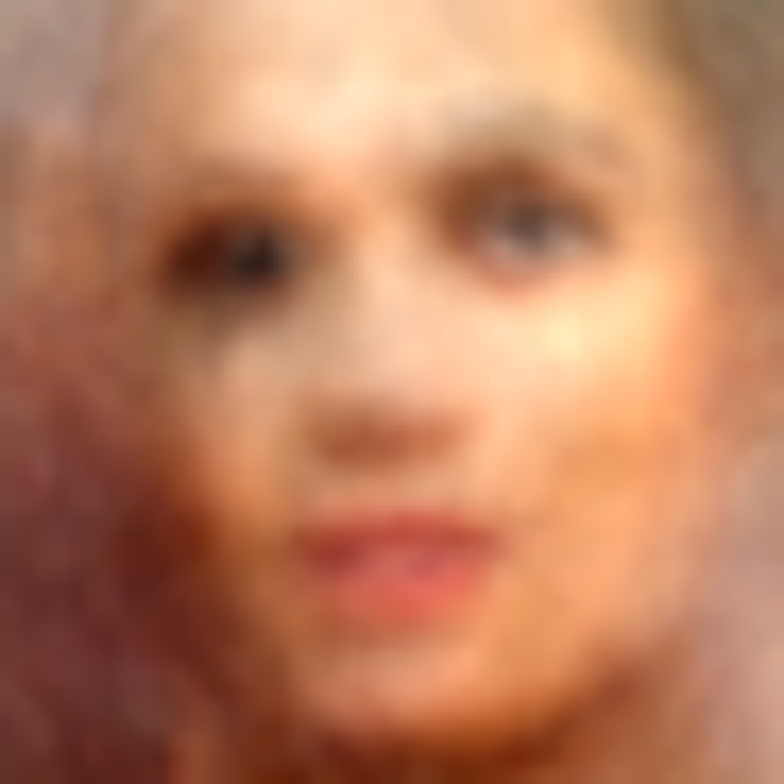} &
    \includegraphics[width=\pwid]{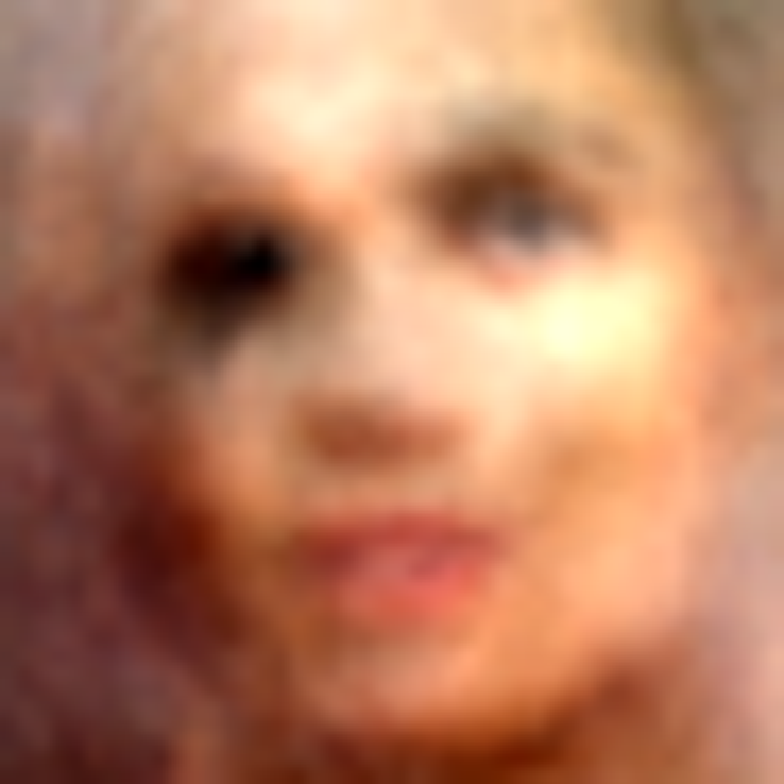} &
    \includegraphics[width=\pwid]{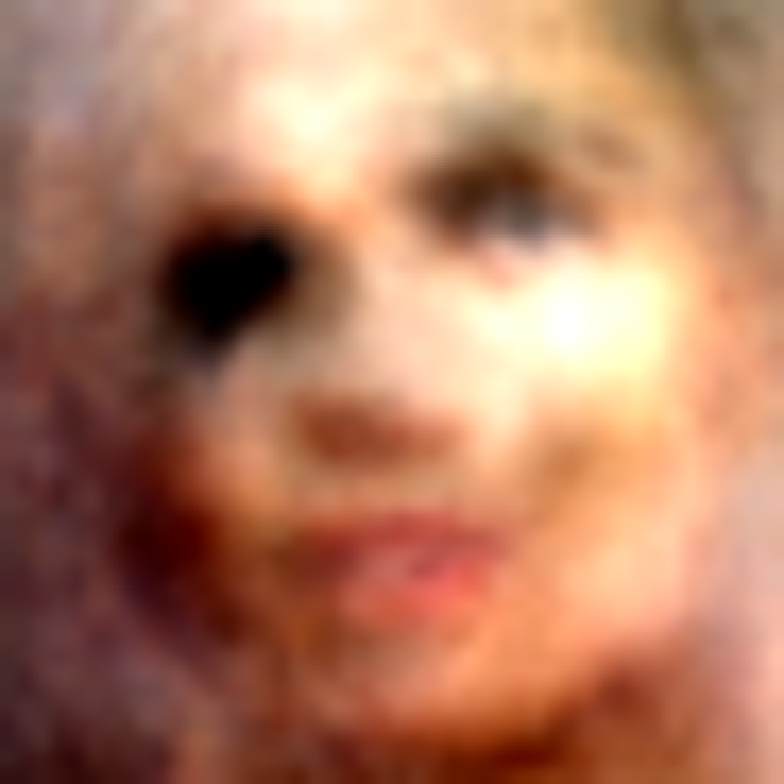}\\[5pt]
                        \multicolumn{3}{l}{\small head down} &
            \multicolumn{1}{c}{$\longrightarrow$} &
            \multicolumn{3}{r}{\small head up}\\
    \includegraphics[width=\pwid]{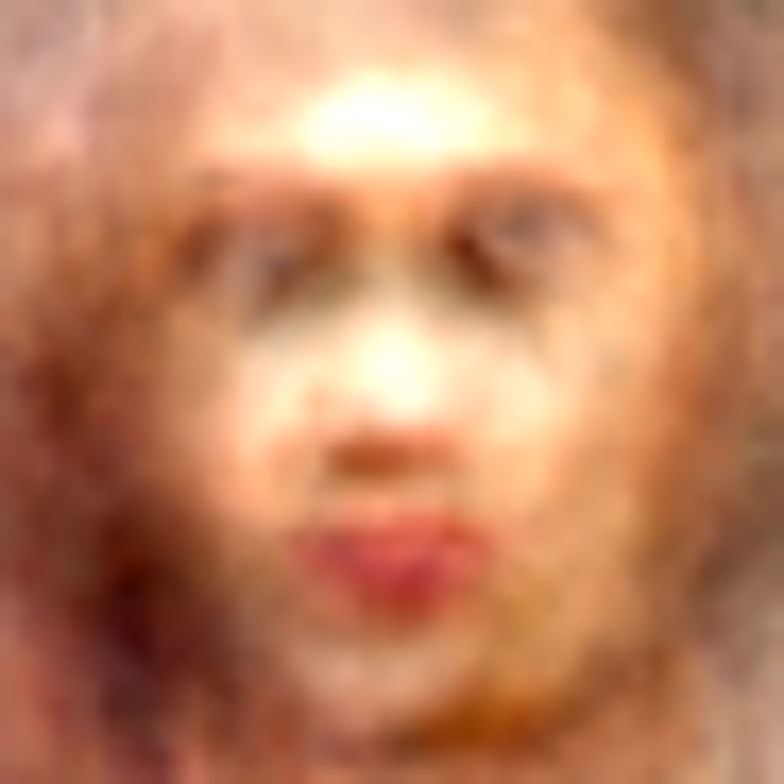} &
    \includegraphics[width=\pwid]{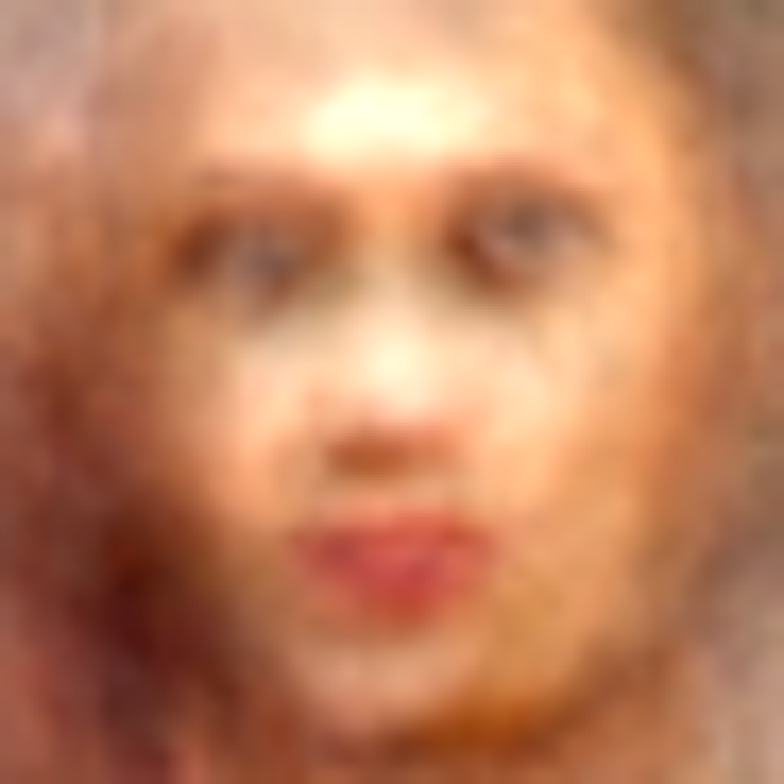} &
    \includegraphics[width=\pwid]{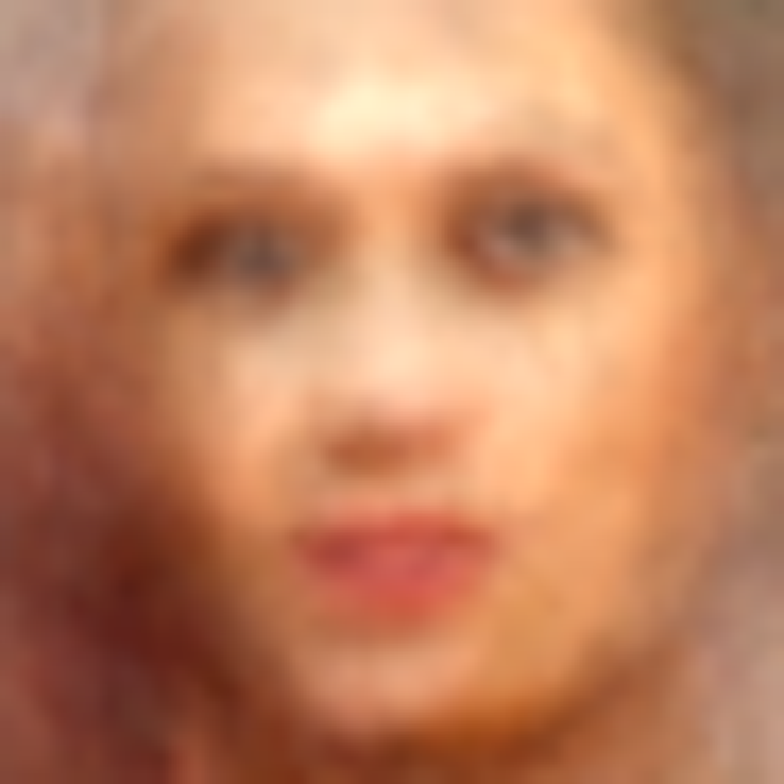} &
    \includegraphics[width=\pwid]{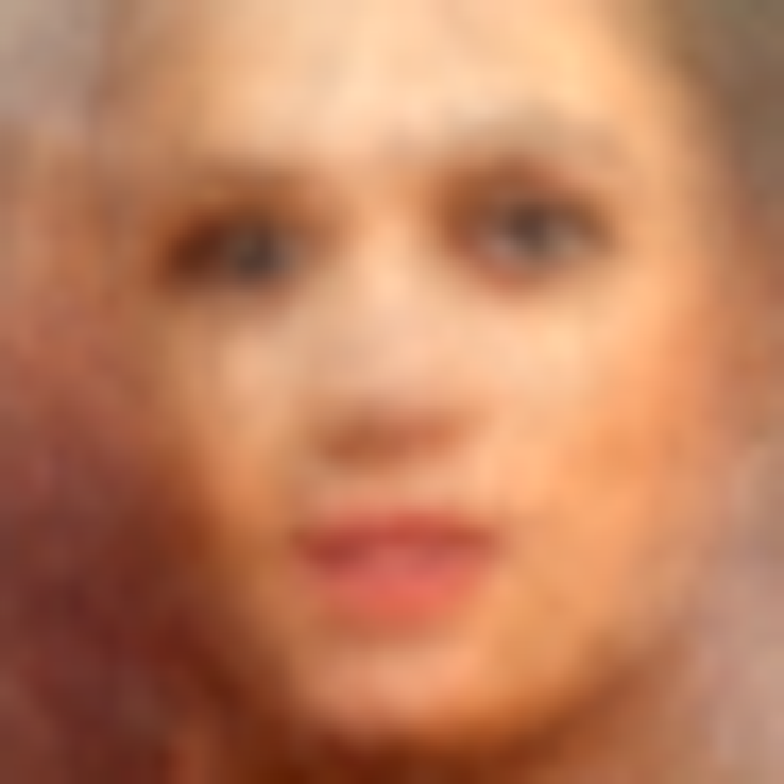} &
    \includegraphics[width=\pwid]{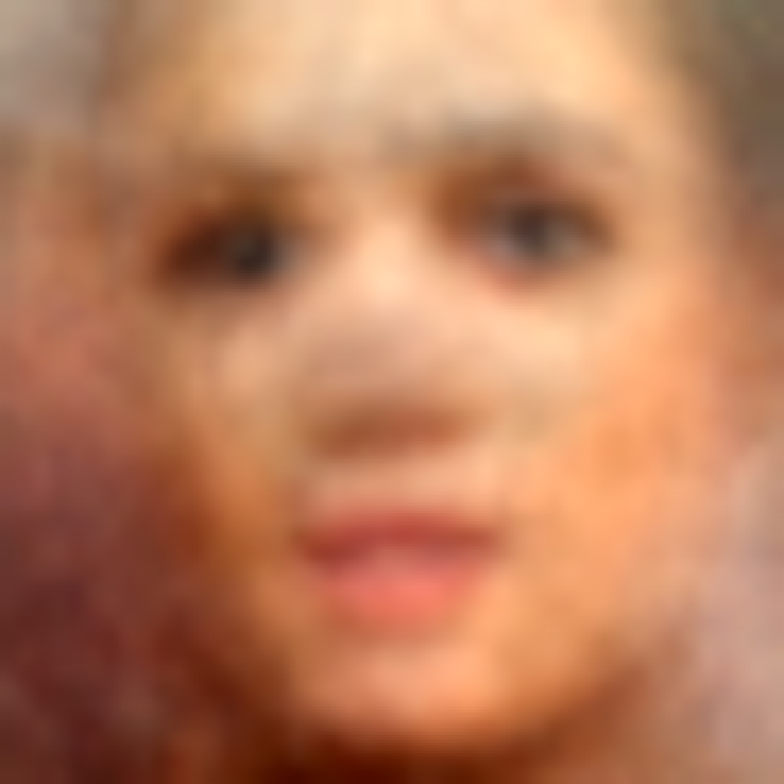} &
    \includegraphics[width=\pwid]{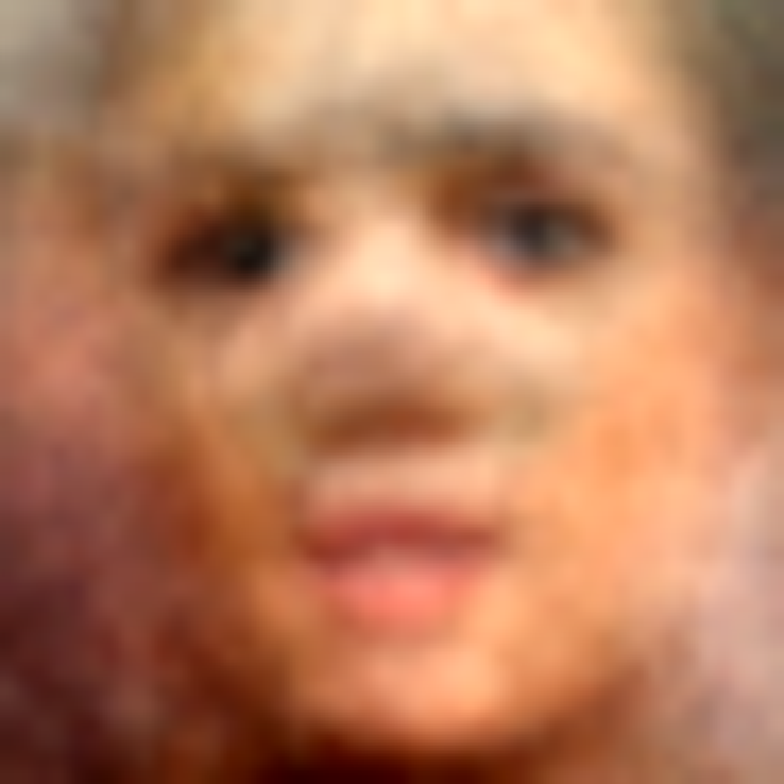} &
    \includegraphics[width=\pwid]{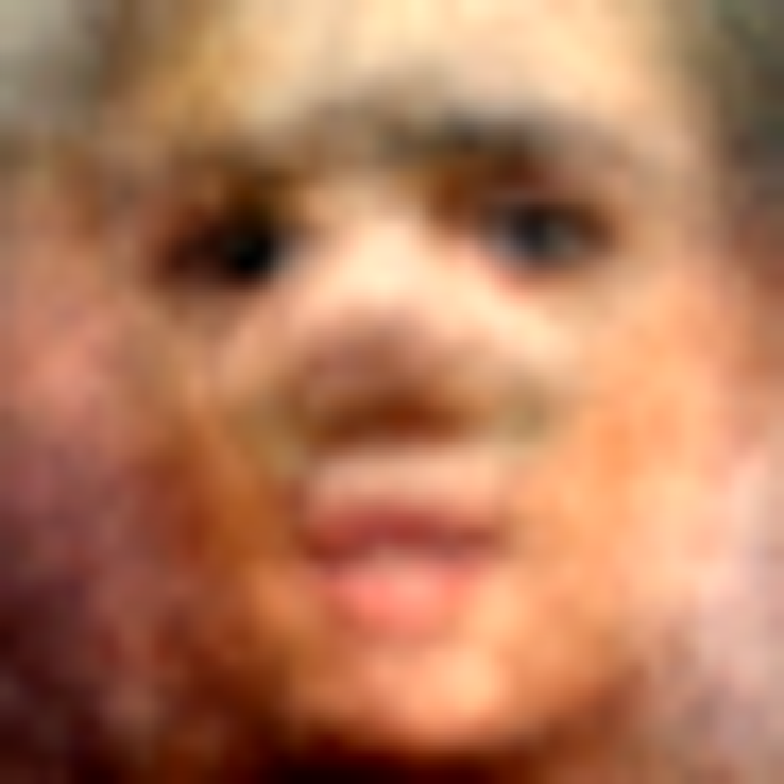}
  \end{tabular}
\vspace{0.2cm}
\caption{\small Semantically altering a face using its
  generic \CNN{} representation and semantically meaningful
  directions in the representation space. Given an image of a face and
  its \CNN{} representation each row above shows the effect of
  altering the face's representation by moving in one direction of the
  representation space and then regressing from the resulting \CNN{}
  representation back to its RGB representation. Each direction was
  learned from labeled training data and corresponds to a specific
  semantic concept: gender, glasses/no-glasses, head angle and head
  tilt. For gender we can see that the left-most image which
  corresponds to male has dark patches corresponding to a beard while
  the right-most image is clearly female. The glasses/no-glasses
  clearly alters the region around the eyes. The last two rows show
  variations caused by changes in head pose. 
  Both the head angle and the tilt (last
  row) are clearly visible.}  \label{fig:AFLW_regenerated}
\end{figure}

\section{Semantic directions in representation space}
\label{sec:semantic}

The \CNN{} is trained with one objective in mind, to learn a
representation where every pair of classes is linearly separable.  The
representation space is thus carved into different volumes
corresponding to the different classes. The results of the paper so
far show each class volume retains significant intra-class
variations. In this section we make a first step towards understanding
how these variations are structured. To proceed we learn a separate
linear regressor from the \rep{} to each of the following variates for
the LFW data-set
\textit{gender}, \textit{have-glasses}, and \textit{pose}. Each 
linear regressor specifies a direction in the representation space
along which a semantic concept varies.

What happens with the images if we alter the representation along this
direction? Can we change the gender or add glasses or continuously
change the pose of the face in the image? We can achieve this if we
extrapolate along an identified direction and then regress back to the
RGB image as described in section \ref{sec:rgb_reconstruction}. In
more detail: the \CNN{} representation, $\bff$, of a face can be
written in terms of its projection onto a semantic direction, such as
gender $\bw_{\text{\tiny gender}}$, found via linear regression and
its component orthogonal to $\bw_{\text{\tiny gender}}$
\begin{align}
        \bff = (\bw_{\text{\tiny gender}}^T \bff)\, \bw_{\text{\tiny
        gender}} + \tilde{\bff}
\end{align}
Then we can create a new \CNN{} representation where the gender
attribute of the face has been altered but not the other factors in
the following simplistic manner.
\begin{align}
    \bff' = \tilde{\bff} + \lambda \bw_{\text{\tiny gender}}
\end{align}
with $\lambda \in [\lambda_{\min}, \lambda_{\max}]$. We then regress
from $\bff'$ back to the RGB image to visualize the result of the
alteration, see figure~\ref{fig:AFLW_regenerated} for some sample
results.

What do the results of our small scale experiment convey? We can see
that altering the pose direction in the \rep{} does correspond well to
the actual image transformation and that changing the gender
corresponds to an altering of the face's color composition. Similarly
glasses/no-glasses alters the appearance of the region around the
eyes. However, it is easy to read too much into the experiments as it
is severely limited by the linear structure of the regressors. And the
fear of hallucinating experimental evidence for the elephant in the
room - the concept of \textit{disentanglement} - means we will leave
our speculations to these comments. However, as such a simple approach
is capable of finding some structures it would be interesting to
investigate if the \rep{} factorizes the variations according to
semantic factors.

%% file: Conclusion.tex
\section{Conclusion}\label{sec:conclusion}
In this paper we have shown that a generic ConvNet representation from
the first fully connected layer retains significant spatial
information.  We demonstrated this fact by solving four different
tasks, that require local spatial information, using the simple common
framework of linear regression from our \CNN{} representation. These
tasks are
\begin{enumerate*}[label=\itshape\roman*\upshape)]
\item 2d facial landmark prediction,
  \item 2d object keypoints
    prediction,
  \item estimation of the RGB values of the original input
    image, and
  \item semantic segmentation, \ie recovering the semantic
label of each pixel.
\end{enumerate*}
The results demonstrated throughout all these
tasks, using diverse datasets, show spatial information is implicitly
encoded in the ConvNet representation and can be easily accessed.
This result is surprising because the employed network was not explicitly
trained to keep spatial information and also the first fully connected
layer is a global image descriptor which aggregates and conflates
appearance features, extracted from the convolutional layers, from all
spatial locations in the image.
